\documentclass[11pt]{article}
\usepackage[margin=1in]{geometry}
\usepackage{amsmath}
\usepackage{booktabs}
\usepackage{hyperref}
\usepackage{array}
\usepackage{amssymb}
\usepackage{graphicx}
\usepackage{longtable}
\usepackage{placeins}
\usepackage[numbers]{natbib}

\title{Benchmarking candidate coverage and rejection policy transfer in typed decision models}

\author{
Jiawen Lu$^{1}$ \qquad Tongtong Wu$^{1}$ \\
$^{1}$Monash University \\
\texttt{jluu0116@student.monash.edu} \qquad
\texttt{tongtong.wu@monash.edu}
}
\date{October 2026}
\hypersetup{pdftitle={Benchmarking candidate coverage and rejection policy transfer in typed decision models},pdfauthor={Jiawen Lu and Tongtong Wu},pdfsubject={Reference-label evaluation, policy transfer and missing-answer sources in Laya, Jev and Qwen}}

\begin{document}
\maketitle

\begin{abstract}
Rejection policies must remain useful as candidate sets and tasks change. We compare Laya, Jev and Qwen2.5-7B-Instruct using public reference labels, testing Laya/Jev policy transfer at equal calibration budgets and all three models on artificial omission, natural retrieval misses and public out-of-scope queries. Source calibration often fails to preserve the target operating point. A Jev policy calibrated on DBpedia rejects 69.3\% of covered Emotion test inputs, while an Emotion policy loses detection entirely. Retrieval exposes a different tradeoff: with ten intent candidates, Laya detects 99.0\% of out-of-scope queries but rejects 48.8\% of covered queries. Separating missing-answer sources reveals these costs alongside retrieval coverage. The benchmark provides shared inputs, explicit decision and failure categories, and reproducible scoring to assess rejection policies under the conditions in which they are reused.
Code and benchmark artifacts are available at \url{https://github.com/luckykevvv/Decision_Model_Benchmark}.
\end{abstract}

\section{Introduction}
Typed decision models select among options supplied at request time \cite{almeidaIntroducingSystemOne2026,convaiinnovationsLaya2026,typesafeaiModelsAPIReference}. A retriever or caller can omit the reference answer, requiring rejection. A threshold that works for one candidate configuration may accept missing answers or reject covered inputs after that configuration changes. Evaluating this portability requires measuring both missing-answer detection and false rejection.

Prior work establishes failures on none-of-the-above questions, absent answers and unsupported intents \cite{tamNoneLessRight2025,fengNoneMeasureUncertainty2020,wangWhenNoAnswer2026,vazeOpenSetRecognitionGood2022,zhanOutofDistributionIntentDetection2021,larsonEvaluationDatasetIntent2019}. Jev experiments further show that thresholds chosen on separate development problems can improve rejection within a task \cite{zhongWhenRightAnswer2026}. We examine what happens when those policies cross tasks or candidate configurations. We also distinguish artificial removal of a known label from retrieval failures and queries outside the supported intent inventory: these sources create different populations of missing answers.

Our benchmark compares Laya, Jev and Qwen on shared public classification and intent inputs. Laya/Jev rejection thresholds use 80 calibration pairs per source task; every target is tested against both the source policy and target recalibration at the same label budget. On CLINC150, all three models are evaluated on artificial omission, natural retrieval misses and global out-of-scope (OOS) queries. Selection, score calibration and matched interface controls diagnose the resulting errors. The results show substantial losses of detection or increases in false rejection after transfer, and distinct tradeoffs across missing-answer sources. Frozen inputs and scoring artifacts make these comparisons reproducible.

\section{Benchmark design}
\label{sec:construction}
\begin{figure}[htbp]
\centering
\includegraphics[width=\textwidth]{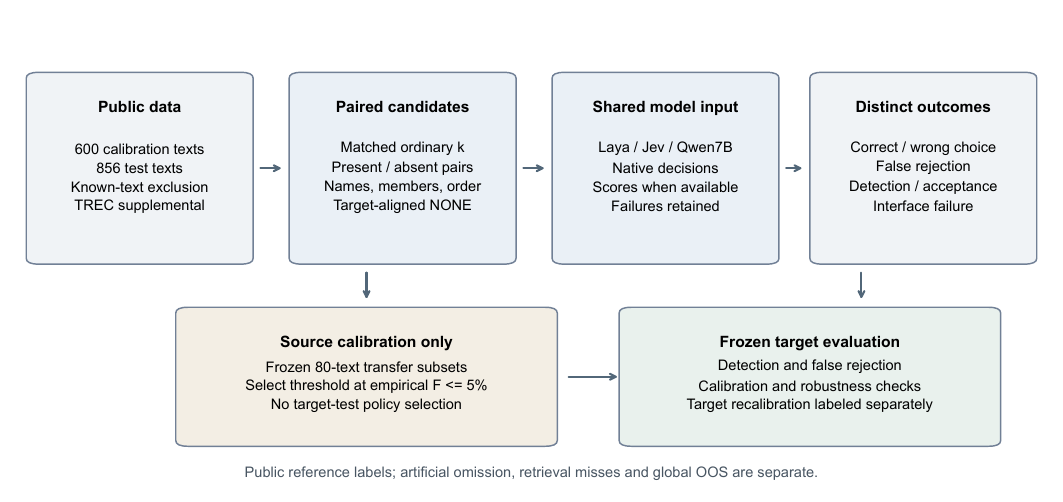}
\caption{Evaluation protocol. Matched candidate sets define covered/missing conditions. Models receive shared inputs, with decisions recorded as correct, wrong, rejected or failed. Source calibration precedes target evaluation; target recalibration uses an equally sized target label set. Counts refer to the classification index.}
\label{fig:v2_protocol}
\end{figure}

The benchmark compares native decisions on shared texts and ordered candidates, then tests score-based rejection policies for Laya and Jev (Figure~\ref{fig:v2_protocol}). Public reference labels define coverage.

\subsection{Reference coverage and decision outcomes}
\label{sec:targets}
For text $x$, ordinary candidate set $C$, and reference class $Y$, define
\[
G(x,C)=\mathbf{1}\{Y\in C\}.
\]
Thus $G=1$ means the reference label is offered. Public OOS queries have $G=0$ and are reported separately.

Let $R(x,C)$ denote rejection, and $\hat c\in C$ an ordinary model choice. Report
\[
D=P(R=1\mid G=0),\qquad F=P(R=1\mid G=1),
\]
alongside correct selection and accepted errors. Native rejection is a valid \texttt{none} response. Covered outcomes are correct, wrong, rejected or failed; missing outcomes are rejected, accepted or failed. Native rates retain all attempted texts, including interface failures.

Normalization over $C$ alone does not establish $P(G=1\mid x,C)$. Selection, coverage-score quality and rejection therefore require separate measurements.

\subsection{Data and sample support}
We use AG News topics \cite{zhangCharacterlevelConvolutionalNetworks2015}, DBpedia entity classes \cite{zhangCharacterlevelConvolutionalNetworks2015}, Emotion labels \cite{saraviaCARERContextualizedAffect2018}, TREC answer types \cite{liLearningQuestionClassifiers2002} and CLINC150 intents/OOS queries \cite{larsonEvaluationDatasetIntent2019}. Table~\ref{tab:tasks} reports distinct texts. Known pilot/discovery texts are excluded, and CLINC queries are removed from the retrieval corpus.

\begin{table}[htbp]
\centering\small
\begin{tabular}{lrrrl}
\toprule
Task & Classes & Cal. texts & Test texts & Ordinary $k$\\
\midrule
AG News & 4 & 80 & 120 & 2, 3\\
DBpedia & 14 & 280 & 420 & 2, 4, 13\\
Emotion & 6 & 120 & 166 & 2, 4, 5\\
TREC* & 6 & 120 & 150 & 2, 4, 5\\
CLINC150 & 150 & 350 & 400 & 2, 5, 10\\
\bottomrule
\end{tabular}
\caption{Sample support from official training/test sources. Classification sampling targets 20 calibration and 30 test texts per class; Emotion surprise has 16 test texts. CLINC includes 300 in-scope queries per split plus 50/100 calibration/test OOS queries. $k$ excludes \texttt{none}. *TREC lacks ABBR test examples and is supplemental.}
\label{tab:tasks}
\end{table}

The study contains 1,456 classification texts and 750 CLINC queries. Interface controls reuse 120 classification texts, with ten calibration and twenty test texts per task. Candidate variants generate 224,964 neural-model responses; the BM25 reference adds 6,900 decisions. The earlier 889-text discovery study is reported separately in Appendix~\ref{sec:discovery_results}.

\subsection{Three missing-answer sources}
\paragraph{Artificial omission.} A wrong-label menu $C_0$ is paired with $C_1$ by substituting the reference label at the same candidate count. Text, instruction and shared descriptions remain fixed. CLINC uses matched $k=5$ menus constructed from ranked wrong intents.

\paragraph{Natural retrieval misses.} A fixed BM25 retriever \cite{robertsonProbabilisticRelevanceFramework2009} supplies top-$k$ intents at $k=2,5,10$. An in-scope query is missing when its reference intent falls outside that menu. Retrieved sets receive no gold-label repair.

\paragraph{Global OOS.} These queries carry the public dataset's label for intents outside its supported 150-intent ontology. They use the same retriever. Each missing-answer source has a separate detection denominator; false rejection uses covered queries.

\FloatBarrier

\section{Results}
\label{sec:results}
\subsection{Calibration fails to preserve rejection after transfer}
\label{sec:confirmation_transfer}
\begin{figure}[htbp]
\centering
\includegraphics[width=\textwidth]{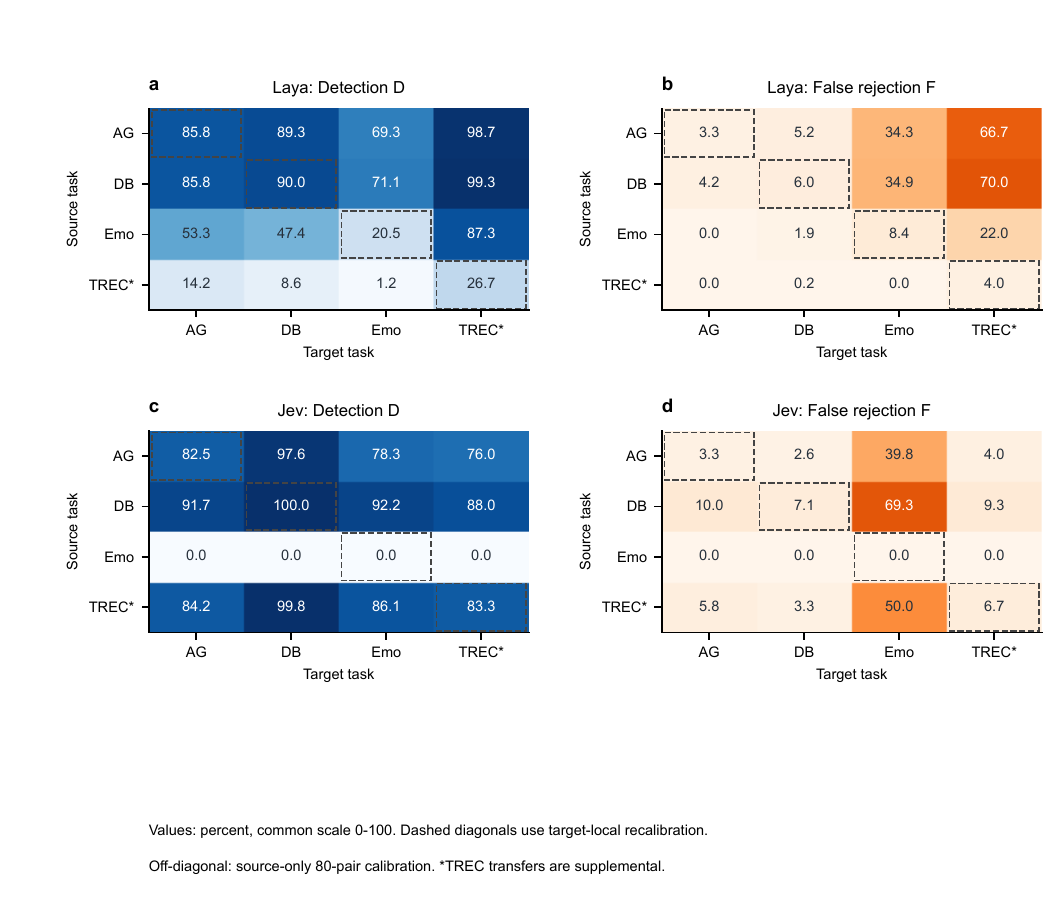}
\caption{Policy transfer at $k=2$, natural names and 80 calibration pairs per task. Rows are sources and columns targets. Off-diagonal cells use source thresholds; dashed diagonals use target labels for recalibration. Target test-text $n=120/420/166/150$ for AG News/DBpedia/Emotion/TREC. Both metrics span 0--100\%. Rates hold thresholds fixed; source data retain marginal text-bootstrap and refitted-source intervals. *TREC is supplemental.}
\label{fig:v2_transfer}
\end{figure}

With 80 calibration pairs per source task, none-score policies exceed 5\% target false rejection in six of twelve task transfers for each of Laya and Jev (Figure~\ref{fig:v2_transfer}). Excluding supplemental TREC leaves three of six transfers above 5\% for each model. The matrix fixes two ordinary candidates and natural names; transfer combinations share test texts.

Jev's DBpedia threshold ($\tau=0.06$) yields $D=92.2\%$ (95\% text-bootstrap interval 88.0--95.8\%) on Emotion, but also $F=69.3\%$ (62.0--76.5\%; $n=166$). Laya's AG News policy similarly gives $D=98.7\%$ and $F=66.7\%$ (58.7--74.0\%; $n=150$) on TREC. High detection after transfer can therefore come at a substantial cost to covered inputs.

Conservative source calibration can lose detection entirely. Jev's Emotion threshold is $\tau=1.00$, which disables rejection under the strict $p(\texttt{none})>\tau$ rule. All targets then have $D/F=0/0$. Recalibrating on 80 AG News pairs restores $D/F=82.5\%/3.3\%$, at the cost of target labels. Even target recalibration can exceed 5\% test false rejection, as both DBpedia diagonal entries show.

\FloatBarrier

\subsection{Retrieval coverage and rejection can move in opposite directions}
\label{sec:missing_sources}
The CLINC150 retriever covers 269/300 in-scope test references at $k=2$, 289/300 at $k=5$ and 295/300 at $k=10$ (Table~\ref{tab:confirmation_intent}). Natural-miss detection thus uses 31, 11 and five queries, respectively; public OOS detection uses a separate 100-query population.

As $k$ increases from two to ten, Laya's OOS detection rises from 83.0\% to 99.0\%, while covered false rejection rises from 15.2\% to 48.8\%. Jev and Qwen retain lower false rejection but lose OOS detection: 84.0\% to 75.0\% and 85.0\% to 63.0\%, respectively. These trajectories combine changes in candidate count, membership and the covered population. Natural-miss estimates at $k=10$ have only five supporting cases.

Artificial omission gives $D/F=86.0\%/27.7\%$ for Laya, $63.7\%/3.3\%$ for Jev and $45.0\%/3.7\%$ for Qwen on 300 matched queries at $k=5$. These intervention rates and the two natural populations describe different failure conditions; each retains its own denominator.

The BM25 gate rejects an artificial menu when its full-ontology winner is absent. Natural menus always contain that winner, so their rejection rates are zero by construction. This fixed-ontology reference exposes a limitation of coupling the rejection gate to the retriever.

\begin{longtable}{llrrrrr}
\caption{Missing-answer sources in CLINC150: native explicit-NONE decisions (\%). Artificial omission uses 300 matched in-scope queries; retrieval uses 300 in-scope and 100 OOS queries. Coverage is the reference-label retrieval hit rate. $D_{\mathrm{miss}}/F_{\mathrm{cov}}$ use the stated missing/covered counts; $D_{\mathrm{OOS}}$ uses 100 queries. Source data retain intervals and failures. *BM25 uses 14,700 labeled utterances and fixed intent identifiers.}\label{tab:confirmation_intent}\\
\toprule Track / $k$ & Model & Coverage & $n_{\mathrm{cov}}/n_{\mathrm{miss}}$ & $D_{\mathrm{miss}}$ & $F_{\mathrm{cov}}$ & $D_{\mathrm{OOS}}$\\\midrule\endfirsthead
\toprule Track / $k$ & Model & Coverage & $n_{\mathrm{cov}}/n_{\mathrm{miss}}$ & $D_{\mathrm{miss}}$ & $F_{\mathrm{cov}}$ & $D_{\mathrm{OOS}}$\\\midrule\endhead
Artificial / 5 & Laya & -- & 300/300 & 86.0 & 27.7 & --\\
Artificial / 5 & Jev & -- & 300/300 & 63.7 & 3.3 & --\\
Artificial / 5 & Qwen7B & -- & 300/300 & 45.0 & 3.7 & --\\
Artificial / 5 & BM25 scope gate* & -- & 300/300 & 82.3 & 17.7 & --\\
Retrieved / 2 & Laya & 89.7 & 269/31 & 90.3 & 15.2 & 83.0\\
Retrieved / 2 & Jev & 89.7 & 269/31 & 83.9 & 3.3 & 84.0\\
Retrieved / 2 & Qwen7B & 89.7 & 269/31 & 71.0 & 6.3 & 85.0\\
Retrieved / 2 & BM25 scope gate* & 89.7 & 269/31 & 0.0 & 0.0 & 0.0\\
Retrieved / 5 & Laya & 96.3 & 289/11 & 90.9 & 30.4 & 92.0\\
Retrieved / 5 & Jev & 96.3 & 289/11 & 72.7 & 4.2 & 77.0\\
Retrieved / 5 & Qwen7B & 96.3 & 289/11 & 54.5 & 2.8 & 67.0\\
Retrieved / 5 & BM25 scope gate* & 96.3 & 289/11 & 0.0 & 0.0 & 0.0\\
Retrieved / 10 & Laya & 98.3 & 295/5 & 100.0 & 48.8 & 99.0\\
Retrieved / 10 & Jev & 98.3 & 295/5 & 80.0 & 3.1 & 75.0\\
Retrieved / 10 & Qwen7B & 98.3 & 295/5 & 60.0 & 2.4 & 63.0\\
Retrieved / 10 & BM25 scope gate* & 98.3 & 295/5 & 0.0 & 0.0 & 0.0\\
\bottomrule
\end{longtable}

\FloatBarrier

\subsection{Selection and score quality capture different failures}
Similar selection accuracy can conceal different rejection behavior. On covered DBpedia inputs at $k=13$, Jev and Qwen select the reference label on 97.4\% and 96.9\% of cases, but detect only 82.4\% and 25.0\% of missing references. Laya detects 84.8\% while rejecting 29.0\% of covered inputs. Full decision partitions appear in Figure~\ref{fig:v2_comparison} and Table~\ref{tab:confirmation_main}.

Score quality helps explain which local policies are useful. On DBpedia at $k=13$, Jev's none score has AUROC 0.987 (95\% text-bootstrap interval 0.980--0.994) and Brier score 0.068 (0.056--0.081); Laya's values are 0.865 (0.845--0.885) and 0.191 (0.169--0.213). Full-calibration policies yield $D/F=98.8\%/3.6\%$ and $34.0\%/2.4\%$, respectively. Local Emotion policies at $k=5$ detect 25.3\% for Jev and 19.3\% for Laya, with 7.8\% false rejection for both. Reliability curves and all local policies are in Figure~\ref{fig:v2_calibration} and Table~\ref{tab:confirmation_calibration}.

\subsection{Names and NONE wording change detection}
\label{sec:confirmation_robustness}
Matched controls use 20 test texts per task (Table~\ref{tab:confirmation_controls}). Effects average variants within each text before paired bootstrap.

Neutral identifiers reduce Qwen detection by 20.0 percentage points on AG News $k=3$ (95\% interval $-37.5$ to $-2.5$) and by 30.6 points on DBpedia $k=13$ ($-52.8$ to $-11.1$), with zero observed change in false rejection. The DBpedia contrast uses 18 texts after excluding invalid response pairs.

On Emotion $k=5$, task-aligned NONE wording increases Laya detection by 25.0 points (10.0--45.0), with a false-rejection change of 0.0 points ($-15.0$ to 15.0). Jev and Qwen show zero detection change. Interface wording therefore affects the operating point differently across models.
\FloatBarrier

\section{Discussion}
Rejection thresholds depend on their calibration setting. The transfer results extend prior demonstrations of local threshold repair \cite{zhongWhenRightAnswer2026}: a useful local policy can become too permissive or too restrictive on another task. Reporting its source configuration, label budget and target $D/F$ makes that change visible. Target recalibration can improve the operating point, but requires additional labels.

Missing-answer sources should also accompany rejection scores. Artificial pairs measure responses to controlled label removal; natural misses depend on upstream retrieval; OOS queries test the supported inventory's boundary. The CLINC results show that retrieving more reference answers can coincide with rejecting more covered inputs. Measuring retrieval coverage, conditional rejection and selection errors together exposes this cost. The lexical gate illustrates the same dependency structurally: sharing a retriever's winner prevents it from rejecting that retriever's menus.

\subsection{Scope and limitations}
Coverage follows public reference labels. Overlapping descriptions can leave acceptable alternatives after label removal, and approximately balanced samples differ from deployment frequencies. Rare-class support is limited, and model pretraining overlap is unknown.

The study evaluates one checkpoint or version per model. Task changes combine distribution and rubric shifts; candidate changes can alter membership and temperature handling. Results therefore describe the tested interfaces and configurations. The empirical calibration constraint provides no population risk guarantee \cite{wangLECLinearExpectation2026}; uncertainty and the exploratory discovery stage are specified in Methods and the supplement.

Larger natural-miss samples and broader model coverage are the next priorities. The protocol supports these extensions through shared inputs, fixed calibration budgets and separate reporting of each missing-answer source.

\FloatBarrier
\section{Methods}
\label{sec:protocol}
\subsection{Frozen sampling and candidate construction}
We sample pinned training/test sources, selecting tests first and excluding known pilot/discovery texts after case folding and whitespace collapse. Seeds 20261003, 20261006 and 20261007 fix the classification index, control subset and 80-pair calibration subsets. Each classification text receives complete-set controls and covered/missing pairs at distinct counts in $\{2,\min(4,M-1),M-1\}$, where $M$ is taxonomy size. Core conditions use one member/order replicate, two naming modes and plain/explicit-NONE interfaces. Counts are not nested.

Neutral keys preserve descriptions and order. Primary NONE wording follows the task target. Controls use five member and two order seeds: order changes preserve membership, while member changes replace one absent distractor at a fixed slot and preserve the covered set. Legacy wording is compared at member/order zero. The earlier confounded stress generator is excluded; the scored controls retain repeated memberships.

CLINC150 uses the original full source at a pinned commit, with two calibration and two test queries per supported intent plus 50/100 calibration/test OOS queries (seed 20261008). All selected query texts are removed from the training retrieval corpus, leaving 14,700 labeled utterances. BM25 uses lower-case alphanumeric tokens, $k_1=1.2$, $b=0.75$, maximum document score per intent and lexical intent-ID tie-breaking. Natural menus are the first $k$ ranked intents; they never use reference labels for repair. Artificial $k=5$ menus use the five top-ranked wrong intents and reference substitution at slot zero. Descriptions mechanically humanize intent identifiers. Source revisions, exact rubrics and reconstruction details are in Appendix~\ref{app:reproduce}.

\subsection{Model interfaces and failures}
Laya 0.3.21 uses a pinned English checkpoint \cite{convaiinnovationsLaya2026,convaiinnovationsLayaEnglishModel2026} with native temperature handling. Jev responses are pinned and verified as \texttt{jev-1.13.0} \cite{almeidaIntroducingSystemOne2026,typesafeaiModelsAPIReference}. Checkpoints, software versions, token budgets and runtime settings are recorded in Table~\ref{tab:confirmation_models} and Appendix~\ref{app:reproduce}.

Qwen2.5-7B-Instruct \cite{qwenteamQwen25TechnicalReport2025,qwenteamQwen257BInstructModelCheckpoint2024} uses greedy BF16 inference, a 2,048-token input cap and a 32-token output cap. Its fixed instruction requests one offered key. The parser accepts an exact key or JSON string containing it; other output forms fail. Qwen supplies discrete choices only.

Models share frozen case/request files. Qwen records 1,369 interface failures and one truncation in core, 877 interface failures in controls and 129 in the intent extension; Laya/Jev responses are structurally valid. Native rates retain failures in their denominators. Paired controls require valid responses on both sides and report eligible $n$ and excluded comparisons.

The BM25 reference uses the same labeled corpus and full-ontology ranker as retrieval. With NONE available, it rejects when the winner is absent; otherwise it selects the highest-ranked offered intent. It requires fixed intent identifiers and labeled corpus access.

\subsection{Scores and native rejection}
Laya/Jev provide three absence scores: negative maximum and entropy normalized by $\log k$ from plain requests, and raw none probability from explicit-NONE requests. Only entropy renormalizes rounded probabilities. Native choices, option probabilities and Jev's separate confidence field are preserved independently.

AUROC gives ties half credit. Brier scores compare $p(\texttt{none})$ with $1-G$ on balanced artificial pairs with valid probabilities on both sides. Reliability uses ten equal-width bins, omitting empty bins. Conditional ordinary calibration normalizes ordinary probability mass; zero mass is recorded as undefined.

\subsection{Calibration and uncertainty}
For each configuration and score, we maximize calibration detection subject to false rejection at most 5\%. Thresholds are observed calibration scores; ties prefer lower false rejection, then higher threshold. Rejection uses strict $s>\tau$. Transfer fixes $k=2$, natural names and 80 source calibration pairs, reused across models and scores. The source threshold is applied to every target; target-local references use their own 80 labeled pairs. The displayed local diagnostics use the largest incomplete counts and full task calibration sets; all configurations remain in the scored artifacts. Count/name transfers use the same frozen subsets. Selection uses calibration data only.

Accepted score-policy decisions use the highest-probability ordinary candidate from the relevant request, with ties resolved by offered order, including zero-mass rows. CLINC threshold diagnostics use the recorded calibration mixture, including OOS for natural retrieval; test rates remain separate by source. Table~\ref{tab:confirmation_intent} reports native decisions.

The sampling unit is a text/query. We average variants within texts, weight texts equally and compute percentile intervals from 2,000 text-bootstrap draws (base seed 20261005, deterministic per-group seeds). Paired controls preserve both sides of each contrast; CLINC rates condition on their stated populations. Primary transfer intervals hold thresholds fixed; supplements also refit thresholds on resampled source data. Wilson intervals apply to one-observation-per-text rates. Intervals are marginal and unadjusted, conditional on recorded tasks and configurations; transfer rows share texts. Discovery conventions are in Appendix~\ref{app:reproduce}.

\subsection{Precision and response validation}
Probability-sum tolerances are 0.001 for Laya and 0.005 per option for Jev's two-decimal scores. Validation checks bounded values, offered keys, response completeness, hashes and duplicate IDs, preserving native choice/score disagreements. Sequence and precision checks are in Appendix~\ref{sec:validation}.

\subsection{Data and code availability}
The confirmation benchmark code, frozen requests, recorded predictions and scoring outputs are available at \url{https://github.com/luckykevvv/Decision_Model_Benchmark}. Released indices reconstruct dataset texts from pinned sources and verify their hashes. Recorded predictions support offline recomputation without model calls. Source revisions and reconstruction details appear in Appendix~\ref{app:reproduce}.

\FloatBarrier
\appendix
\section{Reproduction details}\label{app:reproduce}
\subsection{Frozen sources and reconstruction}
\begin{description}
\item[AG News] \href{https://huggingface.co/datasets/fancyzhx/ag_news/tree/eb185aade064a813bc0b7f42de02595523103ca4}{\texttt{fancyzhx/ag\_news}}.\\Revision: \texttt{eb185aade064a813bc0b7f42de02595523103ca4}.
\item[Emotion] \href{https://huggingface.co/datasets/dair-ai/emotion/tree/cab853a1dbdf4c42c2b3ef2173804746df8825fe}{\texttt{dair-ai/emotion}}.\\Revision: \texttt{cab853a1dbdf4c42c2b3ef2173804746df8825fe}.
\item[DBpedia] \href{https://huggingface.co/datasets/fancyzhx/dbpedia_14/tree/9abd46cf7fc8b4c64290f26993c540b92aa145ac}{\texttt{fancyzhx/dbpedia\_14}}.\\Revision: \texttt{9abd46cf7fc8b4c64290f26993c540b92aa145ac}.
\item[TREC] \href{https://huggingface.co/datasets/SetFit/TREC-QC/tree/0a34640b3cac3affef6f0abdfb81e3d0ecffcf92}{\texttt{SetFit/TREC-QC}}.\\Revision: \texttt{0a34640b3cac3affef6f0abdfb81e3d0ecffcf92}.
\item[CLINC150] \href{https://github.com/clinc/oos-eval/blob/828f8093932c8fe6ca7936c3d2e52903b1c523de/data/data_full.json}{\texttt{clinc/oos-eval}}, original full dataset \cite{larsonEvaluationDatasetIntent2019}.\\Revision: \texttt{828f8093932c8fe6ca7936c3d2e52903b1c523de}.
\end{description}
Parquet sources use the complete official training and test files. TREC uses separate raw files: \texttt{train\_5500.label} for training and \texttt{TREC\_10.label} for testing. Manifests record exact source URLs, SHA-256 values, and row counts.

Discovery selection uses seed 20261001 plus task index, in task order AG News, Emotion, TREC, DBpedia; test selection precedes calibration. New classification selection uses seed 20261003, controls 20261006, equal-budget calibration 20261007 and CLINC queries 20261008. Normalization is case folding and whitespace collapse. Frozen case identifiers, original row indices/splits, label IDs and per-text hashes support reconstruction from pinned sources. Exact reproduction uses saved cases and requests checked against their hashes, rather than a new sample from a current dataset release.

\subsection{Exact questions and option rubrics}
\paragraph{AG News.} Instruction: ``Which topic best describes this news article?''.
\begin{description}
\item[\textnormal{\texttt{World}}] World events, international relations, and politics.
\item[\textnormal{\texttt{Sports}}] Sports, athletes, teams, and competitions.
\item[\textnormal{\texttt{Business}}] Companies, finance, markets, and the economy.
\item[\textnormal{\texttt{Sci/Tech}}] Science, computers, technology, and gadgets.
\end{description}
\paragraph{DBpedia.} Instruction: ``Which category best describes the main entity in this text?''.
\begin{description}
\item[\textnormal{\texttt{Company}}] A company or business.
\item[\textnormal{\texttt{EducationalInstitution}}] A school or university.
\item[\textnormal{\texttt{Artist}}] An artist or entertainer.
\item[\textnormal{\texttt{Athlete}}] An athlete or sportsperson.
\item[\textnormal{\texttt{OfficeHolder}}] A holder of public office.
\item[\textnormal{\texttt{MeanOfTransportation}}] A vehicle or transport type.
\item[\textnormal{\texttt{Building}}] A building or structure.
\item[\textnormal{\texttt{NaturalPlace}}] A natural geographic feature.
\item[\textnormal{\texttt{Village}}] A village or small settlement.
\item[\textnormal{\texttt{Animal}}] An animal or animal species.
\item[\textnormal{\texttt{Plant}}] A plant or plant species.
\item[\textnormal{\texttt{Album}}] A music album.
\item[\textnormal{\texttt{Film}}] A film or movie.
\item[\textnormal{\texttt{WrittenWork}}] A book or other written work.
\end{description}
\paragraph{Emotion.} Instruction: ``Which emotion is most clearly expressed by this text?''.
\begin{description}
\item[\textnormal{\texttt{sadness}}] Feeling sad, disappointed, or unhappy.
\item[\textnormal{\texttt{joy}}] Feeling happy, pleased, or excited.
\item[\textnormal{\texttt{love}}] Feeling affection, love, or care.
\item[\textnormal{\texttt{anger}}] Feeling angry, irritated, or resentful.
\item[\textnormal{\texttt{fear}}] Feeling afraid, anxious, or worried.
\item[\textnormal{\texttt{surprise}}] Feeling surprised or astonished.
\end{description}
\paragraph{TREC.} Instruction: ``What type of answer does this question ask for?''.
\begin{description}
\item[\textnormal{\texttt{ABBR}}] An abbreviation or its expansion.
\item[\textnormal{\texttt{ENTY}}] An entity, object, or substance.
\item[\textnormal{\texttt{DESC}}] A description, definition, reason, or method.
\item[\textnormal{\texttt{HUM}}] A person or group of people.
\item[\textnormal{\texttt{LOC}}] A place or location.
\item[\textnormal{\texttt{NUM}}] A numeric value, date, or quantity.
\end{description}

Neutral keys follow the class order as zero-padded \texttt{option\_000} identifiers. Discovery uses the legacy NONE description ``None of the listed options correctly describes this text'' and places it last. Confirmation inserts NONE at a seeded slot shared by matched coverage/name/wording conditions. Order controls can change that slot while preserving the candidate members. Primary task-aligned descriptions are:
\begin{description}
\item[AG News] None of the listed topics correctly describes the main topic of this news article.
\item[DBpedia] None of the listed categories correctly describes the main entity in this text.
\item[Emotion] None of the listed emotions best describes the emotion expressed in this text.
\item[TREC] None of the listed answer types matches the type of answer this question asks for.
\item[CLINC150] None of the listed intents describes the intent of this utterance.
\end{description}
Exact strings, criterion order, substitutions and request identifiers are retained in frozen JSONL. CLINC descriptions use ``An utterance with an intent related to'' followed by the identifier with underscores replaced by spaces; its NONE option is last. The comparison does not enumerate independent semantic alternatives.

\subsection{Pinned models, runtime and scoring}
Laya's English revision is \texttt{55cf4c4ebb4ebe31b2550e8bdf3bd21b99753851} \cite{convaiinnovationsLayaEnglishModel2026}. Inference records PyTorch 2.8.0, Transformers 4.56.2, NumPy 2.3.3 and Laya 0.3.21; source preparation uses PyArrow 22.0.0. Effective choice temperatures by total offered-option count are 1.90636 for two options, 1.76015 for three to five, 1.00002 for six to ten and 0.5 for eleven or more. Model weights and sequence construction are unchanged by disabling the incompatible optional attention import. Sequence/head lengths are 512/256, with batch eight. Primary confirmation uses one CPU thread.

Jev uses \texttt{jev-1.13.0}, with response-model verification, per-call UTC timestamps and up to four workers. Confirmation limits chunks to 64 questions; discovery packs each text's question collection. Native choices, probabilities and the separate confidence statistic are retained. Tokens, dates and costs are operational metadata, not intrinsic model-efficiency claims. Qwen uses revision \texttt{a09a35458c702b33eeacc393d103063234e8bc28} \cite{qwenteamQwen257BInstructModelCheckpoint2024}, PyTorch 2.8.0 and Transformers 4.56.2. Its fixed system prompt is: ``You are a classifier. Follow the task and option descriptions. Output only the exact key of one offered option. Do not explain.'' Primary BF16/A40 outputs are separated from the preliminary incompatible-hardware attempt.

Confirmation classification scoring records Python 3.13.11 and NumPy 2.5.3 in metric provenance. Percentile bootstrap uses 2,000 draws with base seed 20261005; SHA-derived per-key seeds stabilize each family. Within-text variants are averaged before text resampling. Repeated source draws retain multiplicity when supplementary thresholds are refit. CLINC uses 2,000 query bootstrap draws for conditional rates. Discovery retains seeds 20261001 (2,000 within-model paired draws) and 20261002 (5,000 cross-model/post-hoc draws). Wilson intervals use $z=1.959963984540054$ only where there is one observation per text. AUROC gives score ties half credit; entropy alone renormalizes rounded probabilities. All threshold decisions use strict $s>\tau$.

\subsection{Released artifacts and provenance}
Benchmark release 0.2.0 contains confirmation input indices, frozen requests, recorded predictions, metrics, runtime summaries, model adapters and scoring code. Text reconstruction verifies pinned source files, per-text hashes and the complete original cases-file bytes, including field order and line endings. Offline scoring uses explicit text indices, with their derived file hashes distinguished from original input fingerprints. Original source terms govern dataset use. Manuscript sources and vector figures are maintained separately from the public benchmark. Earlier discovery results remain separate in the appendix; the pilot, initial stress and hardware engineering outputs are excluded from the primary confirmation matrix. Primary collections retain one outcome per request, including explicit failures.

\section{Implementation and precision audits}
\label{sec:validation}
The first checks below concern the preserved discovery stage. Confirmation additionally verifies all 224,964 neural response identifiers, frozen inputs, complete decision partitions and fixed calibration subsets. Exact frozen-input sequence audits find no recorded option capping or state truncation in the Laya inputs. Native Qwen output-format and truncation failures remain separate. The released package recomputes probability and discrete scores from recorded predictions without new model calls.
An independent plan audit verifies the 889 cases, 23,932 requests, 10,188 pairs, matched descriptions and ordering across name variants, and exclusion of pilot and cross-split duplicate texts. Exact Laya tokenizer/sequence replay for all 15,852 test requests finds no option capping, span collisions, instruction shortening, or state truncation. This rules out those specific length failures for the recorded run, not order effects or inaccessible hosted-model preprocessing.

An eight-probe Jev check compares single-question, packed, and repeated packed calls on fixed preflight texts. Choices do not change in these probes; the maximum reported score difference is 0.01 for single versus packed and 0.02 for packed repeats. This small check supports the chosen implementation but cannot establish determinism or universal packing equivalence.

In the final Jev predictions, 123 distributions have a sum deviating from one by more than 0.001, with maximum deviation 0.01, consistent with two-decimal rounding. One native choice is 0.01 below another reported probability. Neither observation is silently repaired. A sensitivity analysis quantizes Laya scores to two decimals while retaining native choices. The largest AUROC change is 0.0688 for DBpedia $k=13$ natural-name entropy (0.7823 to 0.7134). Renormalizing Jev distributions changes any tested AUROC by at most 0.00042. Score precision therefore matters, especially when comparing close rankings or saturated scores.

The benchmark artifacts retain manifests, model-specific runtime records, adapter/analysis code, response identifiers and every measured configuration. Validation rejects incomplete collections and duplicates. Saved native Jev responses allow recovery after interruption; Qwen resumes its frozen model/hardware configuration. Initial engineering events do not replace recorded primary outcomes. A package check reconciles point estimates independently, while isolated recomputation checks the scoring implementation and saved bootstrap outputs. Credentials and request headers are excluded.

\section{Additional confirmation diagnostics}
\label{app:v2_framework}
These diagnostics accompany the main transfer and missing-source results. Each neural model has 39,168 core, 28,920 control and 6,900 intent responses. Figures and tables below retain native partitions, local probability diagnostics and all matched controls.

\subsection{Recorded configurations and native outcomes}

\begin{table}[htbp]
\centering\small
\caption{Recorded confirmation interfaces and configurations. Qwen7B denotes Qwen2.5-7B-Instruct. Model/checkpoint revisions and software versions are retained in runtime manifests; hosted internal compute is not inferred.}\label{tab:confirmation_models}
\begin{tabular}{p{0.13\linewidth}p{0.31\linewidth}p{0.40\linewidth}}
\toprule Model & Frozen model/interface & Recorded configuration\\\midrule
Laya & Laya 0.3.21, pinned revision & CUDA, 1 CPU thread; 512 sequence / 256 head tokens; batch 8. Native temperature handling.\\
Jev & Jev 1.13.0, hosted Choice & Four API workers; up to 64 questions per call; native choices and rounded two-decimal scores.\\
Qwen7B & Qwen2.5-7B-Instruct, pinned revision \cite{qwenteamQwen257BInstructModelCheckpoint2024} & BF16, greedy; batch 32; input cap 2,048 / output cap 32 tokens; exact-key parser; discrete track.\\
\bottomrule
\end{tabular}
\end{table}

\begin{figure}[htbp]
\centering
\includegraphics[width=\textwidth]{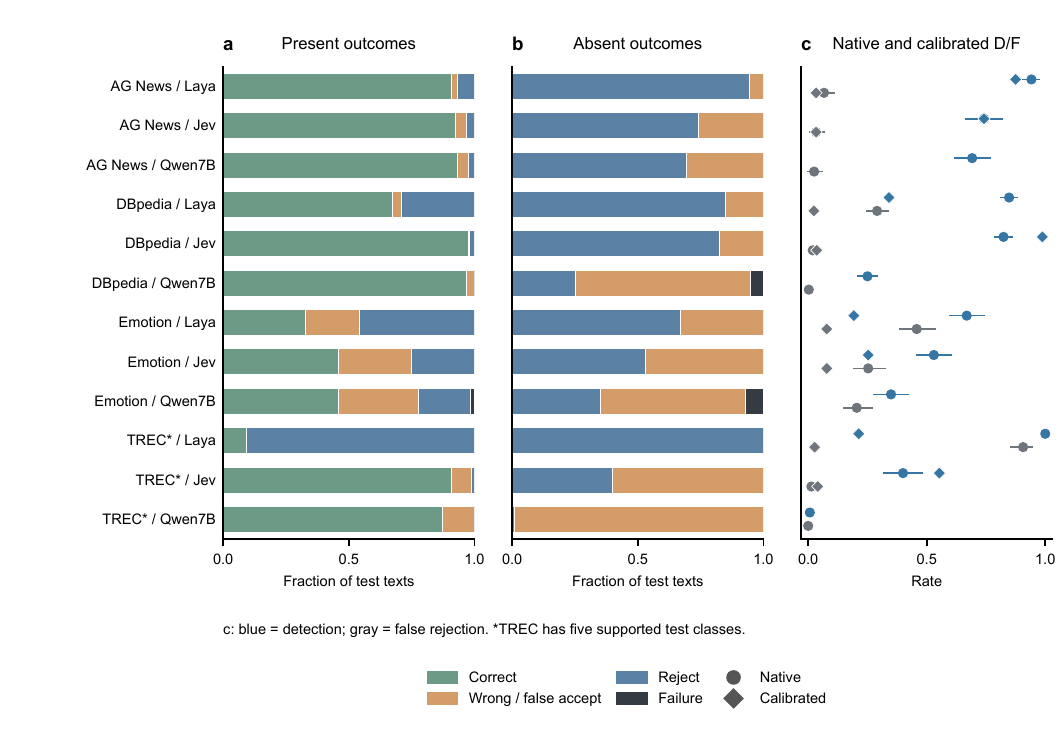}
\caption{Native outcomes and local rejection policies on fresh-text confirmation. Natural names are used at the largest incomplete count: $k=3/13/5/5$ for AG News/DBpedia/Emotion/TREC. Independent test-text $n=120/420/166/150$. a, covered correct, wrong, rejected and failed outcomes. b, missing-reference rejection, acceptance and failure. c, native detection (blue) and false rejection (gray), with marginal 95\% text-bootstrap intervals (2,000 draws); diamonds show none-score policies selected using the full same-task calibration set. Qwen has native points only. TREC is supplemental. Native and score-threshold decisions use their respective recorded interfaces.}
\label{fig:v2_comparison}
\end{figure}
\begin{longtable}{llrrrrrr}
\caption{Reference-label confirmation with natural names at the largest incomplete count. Rates are percentages; covered correct/wrong/rejected/failure and missing rejection/acceptance/failure partitions retain all attempted test texts. Fail $P/A$ denotes covered/missing failure percentages. *TREC has no ABBR test support and is supplemental. Marginal 95\% text-bootstrap intervals are retained in source data.}\label{tab:confirmation_main}\\
\toprule Task & Model & Test $n$ & Correct & Wrong & $F$ & $D$ & Fail $P/A$\\\midrule\endfirsthead
\toprule Task & Model & Test $n$ & Correct & Wrong & $F$ & $D$ & Fail $P/A$\\\midrule\endhead
AG News & Laya & 120 & 90.8 & 2.5 & 6.7 & 94.2 & 0.0/0.0\\
AG News & Jev & 120 & 92.5 & 4.2 & 3.3 & 74.2 & 0.0/0.0\\
AG News & Qwen7B & 120 & 93.3 & 4.2 & 2.5 & 69.2 & 0.0/0.0\\
DBpedia & Laya & 420 & 67.1 & 3.8 & 29.0 & 84.8 & 0.0/0.0\\
DBpedia & Jev & 420 & 97.4 & 0.7 & 1.9 & 82.4 & 0.0/0.0\\
DBpedia & Qwen7B & 420 & 96.9 & 2.9 & 0.2 & 25.0 & 0.0/5.2\\
Emotion & Laya & 166 & 32.5 & 21.7 & 45.8 & 66.9 & 0.0/0.0\\
Emotion & Jev & 166 & 45.8 & 28.9 & 25.3 & 53.0 & 0.0/0.0\\
Emotion & Qwen7B & 166 & 45.8 & 31.9 & 20.5 & 34.9 & 1.8/7.2\\
TREC* & Laya & 150 & 9.3 & 0.0 & 90.7 & 100.0 & 0.0/0.0\\
TREC* & Jev & 150 & 90.7 & 8.0 & 1.3 & 40.0 & 0.0/0.0\\
TREC* & Qwen7B & 150 & 87.3 & 12.7 & 0.0 & 0.7 & 0.0/0.0\\
\bottomrule
\end{longtable}

\subsection{Coverage-score diagnostics}
\begin{figure}[htbp]
\centering
\includegraphics[width=\textwidth]{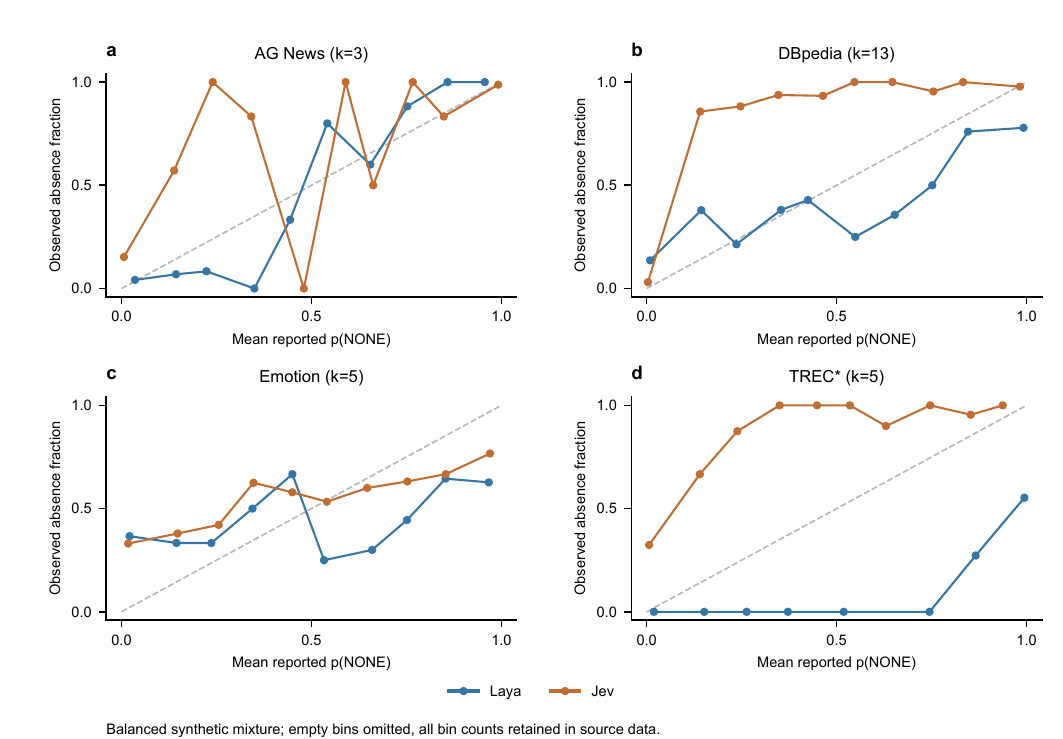}
\caption{Coverage-score reliability varies by task. Raw $p(\texttt{none})$ predicts reference-label absence on a balanced mixture with one covered and one missing response per test text. Tasks, counts and text $n$ match Figure~\ref{fig:v2_comparison}. Ten fixed equal-width bins retain their response counts in source data; empty bins are omitted. Lines are descriptive, and the diagonal denotes agreement between mean predicted probability and observed absence fraction. Only Laya and Jev provide option probabilities. Brier uncertainty uses marginal text bootstrap; paired responses are not independent samples.}
\label{fig:v2_calibration}
\end{figure}
\begin{longtable}{llrrr}
\caption{Coverage-confidence diagnostics on balanced present/absent pairs, using natural names and the largest incomplete count. Brier scores use raw $p(\texttt{none})$; AUROC measures ranking. Laya and Jev provide probability outputs.}\label{tab:confirmation_calibration}\\
\toprule Task & Model & Brier & AUROC & Local calibrated $D/F$ (\%)\\\midrule\endfirsthead
\toprule Task & Model & Brier & AUROC & Local calibrated $D/F$ (\%)\\\midrule\endhead
AG News & Laya & 0.057 & 0.969 & 87.5/3.3\\
AG News & Jev & 0.121 & 0.943 & 74.2/3.3\\
DBpedia & Laya & 0.191 & 0.865 & 34.0/2.4\\
DBpedia & Jev & 0.068 & 0.987 & 98.8/3.6\\
Emotion & Laya & 0.326 & 0.636 & 19.3/7.8\\
Emotion & Jev & 0.277 & 0.704 & 25.3/7.8\\
TREC* & Laya & 0.425 & 0.812 & 21.3/2.7\\
TREC* & Jev & 0.278 & 0.880 & 55.3/4.0\\
\bottomrule
\end{longtable}

These local policies use the full task-specific calibration sample at the largest incomplete count, unlike the $k=2$, 80-pair transfer/local-reference track. Coverage Brier/reliability uses balanced artificial pairs. Conditional ordinary calibration is undefined when the retained ordinary probability mass is zero; the artifact records these counts and any accepted tie-rule decisions.

\subsection{Matched interface contrasts}
\begin{longtable}{lllrp{0.20\linewidth}p{0.20\linewidth}}
\caption{Matched interface controls at the largest incomplete count, using natural names except for the name contrast. Changes in $D/F$ are percentage points with marginal paired-text 95\% bootstrap intervals (2,000 draws). Each task starts with 20 test texts; both-side validity determines eligible $n$. Excluded comparisons remain in source data, while native failure denominators retain all attempts. Order holds members fixed; wording holds keys/order fixed. Variants are not independent texts.}\label{tab:confirmation_controls}\\
\toprule Task & Model & Control & Text $n$ & $\Delta D$ [95\%] & $\Delta F$ [95\%]\\\midrule\endfirsthead
\toprule Task & Model & Control & Text $n$ & $\Delta D$ [95\%] & $\Delta F$ [95\%]\\\midrule\endhead
AG News & Laya & ID minus natural & 20 & -2.5 [-7.5, 0.0] & 12.5 [2.5, 25.0]\\
AG News & Laya & Order 1 minus 0 & 20 & -5.0 [-15.0, 0.0] & -5.0 [-15.0, 0.0]\\
AG News & Laya & Aligned minus old & 20 & 0.0 [0.0, 0.0] & 0.0 [0.0, 0.0]\\
DBpedia & Laya & ID minus natural & 20 & 7.5 [-2.5, 17.5] & 12.5 [0.0, 27.5]\\
DBpedia & Laya & Order 1 minus 0 & 20 & 0.0 [-15.0, 15.0] & -10.0 [-25.0, 0.0]\\
DBpedia & Laya & Aligned minus old & 20 & 15.0 [0.0, 30.0] & 5.0 [0.0, 15.0]\\
Emotion & Laya & ID minus natural & 20 & -12.5 [-30.0, 2.5] & 5.0 [-7.5, 17.5]\\
Emotion & Laya & Order 1 minus 0 & 20 & 5.0 [-10.0, 20.0] & 0.0 [-15.0, 15.0]\\
Emotion & Laya & Aligned minus old & 20 & 25.0 [10.0, 45.0] & 0.0 [-15.0, 15.0]\\
TREC* & Laya & ID minus natural & 20 & 0.0 [0.0, 0.0] & -15.0 [-25.0, -5.0]\\
TREC* & Laya & Order 1 minus 0 & 20 & 0.0 [0.0, 0.0] & 0.0 [-15.0, 15.0]\\
TREC* & Laya & Aligned minus old & 20 & 0.0 [0.0, 0.0] & 15.0 [0.0, 35.0]\\
AG News & Jev & ID minus natural & 20 & 3.5 [0.0, 9.5] & -5.0 [-15.0, 0.0]\\
AG News & Jev & Order 1 minus 0 & 20 & 0.0 [0.0, 0.0] & 0.0 [0.0, 0.0]\\
AG News & Jev & Aligned minus old & 20 & 0.0 [0.0, 0.0] & 0.0 [0.0, 0.0]\\
DBpedia & Jev & ID minus natural & 20 & 1.0 [-3.5, 7.0] & 0.0 [0.0, 0.0]\\
DBpedia & Jev & Order 1 minus 0 & 20 & -4.0 [-11.0, 0.0] & 0.0 [0.0, 0.0]\\
DBpedia & Jev & Aligned minus old & 20 & 5.0 [0.0, 15.0] & 0.0 [0.0, 0.0]\\
Emotion & Jev & ID minus natural & 20 & -8.0 [-23.0, 4.5] & -7.0 [-17.5, 0.5]\\
Emotion & Jev & Order 1 minus 0 & 20 & -1.0 [-15.0, 14.0] & -2.0 [-15.0, 11.0]\\
Emotion & Jev & Aligned minus old & 20 & 0.0 [0.0, 0.0] & 5.0 [0.0, 15.0]\\
TREC* & Jev & ID minus natural & 20 & -11.0 [-25.5, 0.0] & 0.0 [0.0, 0.0]\\
TREC* & Jev & Order 1 minus 0 & 20 & -4.0 [-12.0, 0.0] & 0.0 [0.0, 0.0]\\
TREC* & Jev & Aligned minus old & 20 & 10.0 [0.0, 25.0] & 0.0 [0.0, 0.0]\\
AG News & Qwen7B & ID minus natural & 20 & -20.0 [-37.5, -2.5] & 0.0 [0.0, 0.0]\\
AG News & Qwen7B & Order 1 minus 0 & 20 & 10.0 [0.0, 25.0] & 0.0 [0.0, 0.0]\\
AG News & Qwen7B & Aligned minus old & 20 & 0.0 [0.0, 0.0] & 0.0 [0.0, 0.0]\\
DBpedia & Qwen7B & ID minus natural & 18 & -30.6 [-52.8, -11.1] & 0.0 [0.0, 0.0]\\
DBpedia & Qwen7B & Order 1 minus 0 & 17 & -5.9 [-17.6, 0.0] & 0.0 [0.0, 0.0]\\
DBpedia & Qwen7B & Aligned minus old & 18 & 0.0 [0.0, 0.0] & 0.0 [0.0, 0.0]\\
Emotion & Qwen7B & ID minus natural & 20 & -17.5 [-35.0, -5.0] & 0.0 [0.0, 0.0]\\
Emotion & Qwen7B & Order 1 minus 0 & 19 & -5.3 [-15.8, 0.0] & 0.0 [0.0, 0.0]\\
Emotion & Qwen7B & Aligned minus old & 20 & 0.0 [0.0, 0.0] & 0.0 [0.0, 0.0]\\
TREC* & Qwen7B & ID minus natural & 20 & 0.0 [0.0, 0.0] & 0.0 [0.0, 0.0]\\
TREC* & Qwen7B & Order 1 minus 0 & 20 & 0.0 [0.0, 0.0] & 0.0 [0.0, 0.0]\\
TREC* & Qwen7B & Aligned minus old & 20 & 0.0 [0.0, 0.0] & 0.0 [0.0, 0.0]\\
\bottomrule
\end{longtable}

Each task starts with 20 control test texts. Both-side interface validity determines eligible paired contrasts; excluded comparisons remain recorded. Text-family averaging and paired bootstrap preserve repeated variants without treating them as independent samples.

\subsection{All primary none-score task transfers}

\begin{longtable}{lllp{0.27\linewidth}p{0.27\linewidth}}
\caption{All none-score task transfers at $k=2$ and natural names, with exactly 80 calibration pairs per source/local reference. Cells show target $D/F$ percentages; local references consume target calibration labels. Every ordered task pair is retained. Frozen-threshold and refitted-source uncertainty are recorded separately in source data. *TREC comparisons are supplemental.}\label{tab:confirmation_transfer}\\
\toprule Model & Source & Target & Frozen source $D/F$ & Local reference $D/F$\\\midrule\endfirsthead
\toprule Model & Source & Target & Frozen source $D/F$ & Local reference $D/F$\\\midrule\endhead
Laya & AG News & DBpedia & 89.3/5.2 & 90.0/6.0\\
Laya & AG News & Emotion & 69.3/34.3 & 20.5/8.4\\
Laya & AG News & TREC* & 98.7/66.7 & 26.7/4.0\\
Laya & DBpedia & AG News & 85.8/4.2 & 85.8/3.3\\
Laya & DBpedia & Emotion & 71.1/34.9 & 20.5/8.4\\
Laya & DBpedia & TREC* & 99.3/70.0 & 26.7/4.0\\
Laya & Emotion & AG News & 53.3/0.0 & 85.8/3.3\\
Laya & Emotion & DBpedia & 47.4/1.9 & 90.0/6.0\\
Laya & Emotion & TREC* & 87.3/22.0 & 26.7/4.0\\
Laya & TREC* & AG News & 14.2/0.0 & 85.8/3.3\\
Laya & TREC* & DBpedia & 8.6/0.2 & 90.0/6.0\\
Laya & TREC* & Emotion & 1.2/0.0 & 20.5/8.4\\
Jev & AG News & DBpedia & 97.6/2.6 & 100.0/7.1\\
Jev & AG News & Emotion & 78.3/39.8 & 0.0/0.0\\
Jev & AG News & TREC* & 76.0/4.0 & 83.3/6.7\\
Jev & DBpedia & AG News & 91.7/10.0 & 82.5/3.3\\
Jev & DBpedia & Emotion & 92.2/69.3 & 0.0/0.0\\
Jev & DBpedia & TREC* & 88.0/9.3 & 83.3/6.7\\
Jev & Emotion & AG News & 0.0/0.0 & 82.5/3.3\\
Jev & Emotion & DBpedia & 0.0/0.0 & 100.0/7.1\\
Jev & Emotion & TREC* & 0.0/0.0 & 83.3/6.7\\
Jev & TREC* & AG News & 84.2/5.8 & 82.5/3.3\\
Jev & TREC* & DBpedia & 99.8/3.3 & 100.0/7.1\\
Jev & TREC* & Emotion & 86.1/50.0 & 0.0/0.0\\
\bottomrule
\end{longtable}

Maximum/entropy policies, count/name transfers and source-refit intervals are retained in the scored artifacts rather than selected by target test performance. Qwen has no probability-transfer entries. CLINC's native three-source comparison is reported in the main text; its supplemental score thresholds use the recorded calibration mixture and retain separate in-scope-miss and OOS test outcomes.

\section{Discovery construction and scope}
\label{app:discovery_construction}
This stage is exploratory. The later protocol was frozen before its model calls but was not a registered preregistration; it also changed NONE wording and placement, so differences between stages combine sample and interface changes.
The initial Laya/Jev study contains 889 texts: 300 calibration and 589 test examples. AG News contributes 40/80 calibration/test texts, DBpedia 140/280, Emotion 60/120 and TREC 60/109. Sampling requested ten calibration and twenty test texts per class using seed 20261001, excluding the earlier 300-text pilot and task-specific cross-split normalized duplicates. Only nine TREC ABBR test examples were available; every other represented class supplied twenty.

The discovery construction uses one seeded candidate sample per text/count and non-nested counts. Natural/neutral keys preserve members, descriptions and order, while the reference label is substituted into a wrong-label menu to produce matched present/absent candidates. NONE is last and uses the legacy description ``None of the listed options correctly describes this text.'' Full-set order follows class order, so full-to-pair comparisons can also change order. Each model has 3,556 full-set and 20,376 paired-condition requests, totaling 23,932. The following discovery tables and ROC figure retain this stage's original estimates and uncertainty. They are not estimates from the new confirmation sample.

\section{Discovery results}
\label{sec:discovery_results}
This section preserves the original 889-text Laya/Jev discovery study. Its observations are exploratory and are not pooled with the fresh-text confirmation in Section~\ref{sec:results}.
\subsection{Native decisions have task-dependent tradeoffs}
\begin{table}[htbp]
\centering\small
\setlength{\tabcolsep}{4pt}
\begin{tabular}{llrrrrr}
\toprule
Task ($n$; $k$) & Model & Full acc. & $D$ & $F$ & $D-F$ & AUC$_{\rm none}$ \\
\midrule
AG News (80; 3) & Laya & 95.0 & 87.5 & 5.0 & 82.5 & 0.980 \\
AG News (80; 3) & Jev & 95.0 & 77.5 & 7.5 & 70.0 & 0.931 \\
DBpedia (280; 13) & Laya & 86.4 & 77.1 & 18.6 & 58.6 & 0.868 \\
DBpedia (280; 13) & Jev & 99.3 & 80.7 & 1.4 & 79.3 & 0.992 \\
Emotion (120; 5) & Laya & 51.7 & 68.3 & 45.8 & 22.5 & 0.676 \\
Emotion (120; 5) & Jev & 51.7 & 58.3 & 32.5 & 25.8 & 0.698 \\
TREC (109; 5) & Laya & 87.2 & 97.2 & 69.7 & 27.5 & 0.865 \\
TREC (109; 5) & Jev & 92.7 & 24.8 & 0.0 & 24.8 & 0.846 \\
\bottomrule
\end{tabular}
\caption{Natural-name results at the largest incomplete ordinary candidate count for each task. Full accuracy uses all classes without none. Rates and $D-F$ are percentages/percentage points; none-score AUC compares matched present/absent requests with none. Full uncertainty intervals and all configurations appear in Appendix~\ref{app:all_results}.}
\label{tab:main}
\end{table}

Table~\ref{tab:main} reports natural names at the largest incomplete set for each task; Appendix~\ref{app:all_results} contains every count/name configuration. AG News full-set accuracy is 95.0\% for both models. At $k=3$, Laya has $D=87.5\%$, $F=5.0\%$, while Jev has $D=77.5\%$, $F=7.5\%$. The paired Jev-minus-Laya detection difference is $-10.0$ percentage points (95\% bootstrap interval [$-18.8$, $-2.5$]); the false-rejection difference interval includes zero.

DBpedia differs: full-set accuracy is 86.4\% for Laya and 99.3\% for Jev. At $k=13$, their false-rejection rates are 18.6\% and 1.4\%; the paired difference is $-17.1$ points [$-21.8$, $-12.9$]. The detection difference interval includes zero. Emotion is difficult for both checkpoints under this protocol, with 51.7\% full-set accuracy for both natural-name conditions. Jev reduces false rejection from 45.8\% to 32.5\% but also reduces detection from 68.3\% to 58.3\%. These outcomes do not support a single task-independent model ranking.

\subsection{TREC exposes opposite native rejection regimes}
At five TREC candidates, Laya rejects 97.2\% of absent cases but also 69.7\% of present controls. Jev's rates are 24.8\% and 0/109. Zero observed false rejections has a Wilson upper endpoint of approximately 3.4\%, not a zero-risk guarantee. Full-set accuracy is 87.2\% for Laya and 92.7\% for Jev. A high detection number alone would favor Laya despite frequent rejection of valid requests; a false-rejection-only measure would favor Jev while ignoring missed absence. Their native $D-F$ gaps are only 27.5 and 24.8 points, showing that this scalar also conceals the distinct operating regimes.

\begin{table}[htbp]
\centering\small
\begin{tabular}{lrrrrr}
\toprule
Task & Common $n$ & Laya $D$ & Laya $F$ & Jev $D$ & Jev $F$ \\
\midrule
AG News & 72 & 91.7 & 1.4 & 80.6 & 4.2 \\
DBpedia & 242 & 82.6 & 14.0 & 79.8 & 1.7 \\
Emotion & 47 & 91.5 & 42.6 & 85.1 & 21.3 \\
TREC & 91 & 100.0 & 69.2 & 24.2 & 0.0 \\
\bottomrule
\end{tabular}
\caption{The same texts correctly classified by both models with full natural-name options. Rates are percentages; $k$ matches Table~\ref{tab:main}. This selected subset supplements overall results rather than estimating population performance.}
\label{tab:common}
\end{table}

The contrast remains in a common competence-conditioned subset (Table~\ref{tab:common}). Among 91 TREC texts both models classify correctly with full natural-name options, Laya has $D=100\%$, $F=69.2\%$; Jev has $D=24.2\%$, $F=0\%$. On DBpedia's 242 common-correct texts, false rejection is 14.0\% and 1.7\%. The subset is selected by both models and supplements overall rates rather than defining a deployment population.

\subsection{Ranking does not determine the operating point}
\begin{table}[htbp]
\centering\small
\begin{tabular}{llrrrr}
\toprule
Task & Model & Cal. $n$ & None threshold & Test $D$ & Test $F$ \\
\midrule
AG News & Laya & 40 & 0.4756 & 86.3 & 5.0 \\
AG News & Jev & 40 & 0.8800 & 70.0 & 1.3 \\
DBpedia & Laya & 140 & 0.9999 & 23.9 & 2.1 \\
DBpedia & Jev & 140 & 0.0600 & 98.9 & 3.2 \\
Emotion & Laya & 60 & 0.9846 & 18.3 & 5.8 \\
Emotion & Jev & 60 & 0.9800 & 12.5 & 4.2 \\
TREC & Laya & 60 & 0.9929 & 33.9 & 3.7 \\
TREC & Jev & 60 & 0.1400 & 45.0 & 1.8 \\
\bottomrule
\end{tabular}
\caption{Thresholds maximize calibration detection subject to empirical calibration false rejection at most 5\%, with frozen test evaluation. Natural names and $k$ match Table~\ref{tab:main}; reject iff $p(\texttt{none})>t$. Test rates are percentages, not a guaranteed risk bound.}
\label{tab:calibration}
\end{table}

\begin{figure}[htbp]
\centering
\includegraphics[width=\textwidth]{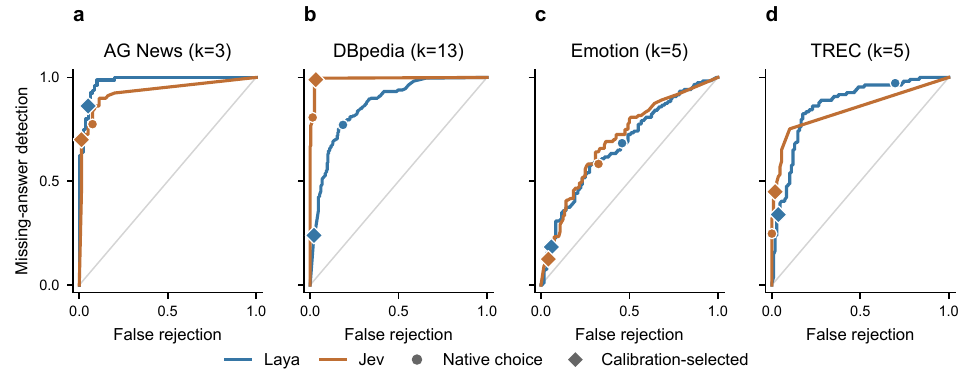}
\caption{Discovery none-score detection--false-rejection curves on matched test pairs with natural names and $k$ from Table~\ref{tab:main}. Colors identify models; circles show native none choices and diamonds calibration-only selected thresholds. Native choices need not lie on a scalar none-threshold curve. Lines are empirical test diagnostics, with no test-selected recommended threshold or uncertainty band; Table~\ref{tab:main} and the appendix provide uncertainty.}
\label{fig:roc}
\end{figure}
Laya's TREC none-score AUROC is 0.865, yet its calibration-selected none threshold reduces detection to 33.9\% while bringing false rejection to 3.7\%. Jev's AUROC is 0.846; calibration increases detection to 45.0\% with false rejection 1.8\%. Thus similar ranking quality can accompany very different native choices and different calibrated tradeoffs (Figure~\ref{fig:roc}).

DBpedia Jev has none-score AUROC 0.992 and a test operating point of $D=98.9\%$, $F=3.2\%$, versus Laya's $D=23.9\%$, $F=2.1\%$. On Emotion, conservative thresholds leave low detection: 18.3\% for Laya and 12.5\% for Jev, with false rejection 5.8\% and 4.2\%. The nominal calibration target does not ensure test $F\leq5\%$.

\subsection{Names and count are meaningful interface controls}
For Laya TREC $k=5$, neutral keys reduce false rejection from 69.7\% to 40.4\%, a paired reduction of 29.4 points [20.2, 38.5]. Descriptions, candidate members, and order match. However, names are not a universal remedy: both models still have substantial Emotion errors, and effects vary across tasks. TREC's abbreviated names and generic NONE wording limit interpretation of this discovery contrast.

For Laya DBpedia, native detection falls from 93.2\% at $k=2$ to 77.1\% at $k=13$, while false rejection rises from 10.7\% to 18.6\%. Candidate composition, position, and effective temperature also vary, so the observation is conditional on this design, not a pure causal size effect. Jev's corresponding native rates change from 97.9\%/2.9\% to 80.7\%/1.4\%. All intermediate configurations and score results remain in the appendix.

\section{Discovery analysis of policy transfer}
\label{sec:transfer}
The preceding calibration results use labels from the same task, candidate count, and naming condition as the test configuration. To examine portability, we additionally freeze each source-calibrated threshold and apply it to every other recorded configuration of the same model. All three scores and all ordered source--target combinations are retained. Target-local recalibration is a separate reference requiring target calibration labels; no target test outcome selects the source, score, or threshold. This analysis was added after inspecting the discovery data and is not an independent confirmation experiment.

Table~\ref{tab:transfer} reports all cross-task none-score transfers at two ordinary candidates with natural names. Laya's AG News threshold yields $D=97.2\%$, $F=67.9\%$ on TREC; target-local calibration yields $D=57.8\%$, $F=9.2\%$. Jev's DBpedia threshold produces $D=92.5\%$, $F=68.3\%$ on Emotion, whereas target-local calibration selects threshold 1.00 and obtains zero detection and zero false rejection. That same Emotion threshold rejects no samples on any other task because rejection uses a strict greater-than comparison. Avoiding false rejection can therefore coincide with disabling absence detection.

Across the 40 recorded transfers that change only the task, observed target $F$ exceeds 5\% in 22 configurations for Laya and 18 for Jev using the none score. These counts describe correlated configurations, not independent experimental samples, deployment prevalence, or significance tests. Even target-local calibration can exceed the empirical 5\% budget on test data. The results support reporting both portability failures and detection losses rather than interpreting source calibration as a universal rejection policy.

The analysis artifacts also retain count-only, name-only and mixed transfers, marginal Wilson intervals, and error decomposition. Score-based policies accept the highest-scoring ordinary candidate from the relevant request, with ties resolved by frozen offered order; their accepted errors are reported separately from native choices. For the none score this uses the explicit-none request, while maximum and entropy use plain requests. Cross-task changes here do not establish held-out domain or new-schema generalization. Section~\ref{sec:confirmation_transfer} reports fresh-text fixed-budget transfer; Appendix~\ref{app:v2_framework} retains complete confirmation tables.

A score audit finds zero total ordinary probability mass in 6 Laya and 49 Jev present-test explicit-none responses, out of 3,374 configuration responses per model. These are correlated response counts rather than independent text counts. If a threshold accepts such a response, the ordinary argmax follows the offered-order tie rule, so its choice is not supported by a positive ordinary score. Conditional ordinary calibration is undefined for these rows; their decision outcomes remain in the recorded denominators.

\subsection{Discovery cross-task transfer table}
\begin{longtable}{lllrrrr}
\caption{Post-hoc cross-task transfer at $k=2$ with natural names and the none score. Thresholds use source calibration only. Local columns use target calibration labels. All ordered task transfers are reported; rates are percentages. These are discovery results, with marginal Wilson intervals in the accompanying JSON.}\label{tab:transfer}\\
\toprule Model & Source & Target & $D$ & $F$ & Local $D$ & Local $F$\\\midrule\endfirsthead
\toprule Model & Source & Target & $D$ & $F$ & Local $D$ & Local $F$\\\midrule\endhead
Laya & DBpedia & AG News & 81.2 & 2.5 & 91.2 & 6.2\\
Laya & Emotion & AG News & 63.7 & 2.5 & 91.2 & 6.2\\
Laya & TREC & AG News & 62.5 & 1.2 & 91.2 & 6.2\\
Laya & AG News & DBpedia & 93.9 & 11.8 & 84.3 & 2.1\\
Laya & Emotion & DBpedia & 56.8 & 0.7 & 84.3 & 2.1\\
Laya & TREC & DBpedia & 54.3 & 0.7 & 84.3 & 2.1\\
Laya & AG News & Emotion & 80.8 & 50.0 & 29.2 & 9.2\\
Laya & DBpedia & Emotion & 57.5 & 19.2 & 29.2 & 9.2\\
Laya & TREC & Emotion & 26.7 & 8.3 & 29.2 & 9.2\\
Laya & AG News & TREC & 97.2 & 67.9 & 57.8 & 9.2\\
Laya & DBpedia & TREC & 90.8 & 23.9 & 57.8 & 9.2\\
Laya & Emotion & TREC & 62.4 & 10.1 & 57.8 & 9.2\\
Jev & DBpedia & AG News & 88.8 & 8.8 & 78.8 & 3.8\\
Jev & Emotion & AG News & 0.0 & 0.0 & 78.8 & 3.8\\
Jev & TREC & AG News & 83.8 & 7.5 & 78.8 & 3.8\\
Jev & AG News & DBpedia & 95.0 & 0.0 & 100.0 & 3.6\\
Jev & Emotion & DBpedia & 0.0 & 0.0 & 100.0 & 3.6\\
Jev & TREC & DBpedia & 97.5 & 2.5 & 100.0 & 3.6\\
Jev & AG News & Emotion & 68.3 & 33.3 & 0.0 & 0.0\\
Jev & DBpedia & Emotion & 92.5 & 68.3 & 0.0 & 0.0\\
Jev & TREC & Emotion & 85.0 & 53.3 & 0.0 & 0.0\\
Jev & AG News & TREC & 50.5 & 0.0 & 70.6 & 1.8\\
Jev & DBpedia & TREC & 80.7 & 5.5 & 70.6 & 1.8\\
Jev & Emotion & TREC & 0.0 & 0.0 & 70.6 & 1.8\\
\bottomrule
\end{longtable}

\section{Complete discovery configurations}\label{app:all_results}
The following tables retain both naming modes and all ordinary candidate counts. Task-specific test sample sizes are AG News 80, DBpedia 280, Emotion 120, and TREC 109; these apply to every corresponding configuration, not to each response as an independent sample. Nat. denotes natural names; ID denotes neutral keys. All within-model intervals are marginal, conditional on the frozen sample, candidate draw, prompts, and runtime.
\begingroup\footnotesize\setlength{\tabcolsep}{3pt}
\begin{longtable}{lllrr}
\caption{Full-candidate test accuracy with Wilson 95\% intervals, percentages.}\label{tab:all_baselines}\\
\toprule
Task & Names & Model & Without none [95\%] & With none [95\%]\\
\midrule\endfirsthead
\toprule
Task & Names & Model & Without none [95\%] & With none [95\%]\\
\midrule\endhead
AG News & natural & Laya & 95.0 [87.8, 98.0] & 90.0 [81.5, 94.8] \\
AG News & natural & Jev & 95.0 [87.8, 98.0] & 90.0 [81.5, 94.8] \\
AG News & neutral & Laya & 95.0 [87.8, 98.0] & 90.0 [81.5, 94.8] \\
AG News & neutral & Jev & 96.3 [89.5, 98.7] & 90.0 [81.5, 94.8] \\
DBpedia & natural & Laya & 86.4 [81.9, 90.0] & 77.1 [71.9, 81.7] \\
DBpedia & natural & Jev & 99.3 [97.4, 99.8] & 97.5 [94.9, 98.8] \\
DBpedia & neutral & Laya & 88.2 [83.9, 91.5] & 75.0 [69.6, 79.7] \\
DBpedia & neutral & Jev & 98.9 [96.9, 99.6] & 98.2 [95.9, 99.2] \\
Emotion & natural & Laya & 51.7 [42.8, 60.4] & 33.3 [25.5, 42.2] \\
Emotion & natural & Jev & 51.7 [42.8, 60.4] & 45.8 [37.2, 54.7] \\
Emotion & neutral & Laya & 50.0 [41.2, 58.8] & 37.5 [29.4, 46.4] \\
Emotion & neutral & Jev & 53.3 [44.4, 62.0] & 46.7 [38.0, 55.6] \\
TREC & natural & Laya & 87.2 [79.6, 92.2] & 26.6 [19.2, 35.6] \\
TREC & natural & Jev & 92.7 [86.2, 96.2] & 92.7 [86.2, 96.2] \\
TREC & neutral & Laya & 83.5 [75.4, 89.3] & 51.4 [42.1, 60.6] \\
TREC & neutral & Jev & 89.0 [81.7, 93.6] & 89.0 [81.7, 93.6] \\
\bottomrule
\end{longtable}
\endgroup

\begingroup\footnotesize\setlength{\tabcolsep}{3pt}
\begin{longtable}{llrrr}
\caption{All paired native-choice rates. D/F include Wilson 95\% intervals; D-F uses 2,000 paired-text bootstrap draws. Rates are percentages and differences percentage points. Nat.=natural names; ID=neutral keys.}\label{tab:all_rates}\\
\toprule
Group ($k$, names) & Model & $D$ [95\%] & $F$ [95\%] & $D-F$ [95\%]\\
\midrule\endfirsthead
\toprule
Group ($k$, names) & Model & $D$ [95\%] & $F$ [95\%] & $D-F$ [95\%]\\
\midrule\endhead
AG News 2 Nat. & Laya & 88.8 [80.0, 94.0] & 6.3 [2.7, 13.8] & 82.5 [73.8, 91.3] \\
AG News 2 Nat. & Jev & 83.8 [74.2, 90.3] & 8.8 [4.3, 17.0] & 75.0 [65.0, 83.8] \\
AG News 2 ID & Laya & 91.3 [83.0, 95.7] & 5.0 [2.0, 12.2] & 86.3 [77.5, 93.8] \\
AG News 2 ID & Jev & 85.0 [75.6, 91.2] & 8.8 [4.3, 17.0] & 76.3 [66.3, 85.0] \\
AG News 3 Nat. & Laya & 87.5 [78.5, 93.1] & 5.0 [2.0, 12.2] & 82.5 [73.8, 91.3] \\
AG News 3 Nat. & Jev & 77.5 [67.2, 85.3] & 7.5 [3.5, 15.4] & 70.0 [60.0, 80.0] \\
AG News 3 ID & Laya & 87.5 [78.5, 93.1] & 7.5 [3.5, 15.4] & 80.0 [70.0, 90.0] \\
AG News 3 ID & Jev & 80.0 [70.0, 87.3] & 8.8 [4.3, 17.0] & 71.3 [61.3, 81.3] \\
DBpedia 2 Nat. & Laya & 93.2 [89.6, 95.6] & 10.7 [7.6, 14.9] & 82.5 [77.9, 87.1] \\
DBpedia 2 Nat. & Jev & 97.9 [95.4, 99.0] & 2.9 [1.5, 5.5] & 95.0 [92.1, 97.5] \\
DBpedia 2 ID & Laya & 92.5 [88.8, 95.0] & 7.9 [5.2, 11.6] & 84.6 [80.0, 88.9] \\
DBpedia 2 ID & Jev & 98.2 [95.9, 99.2] & 2.9 [1.5, 5.5] & 95.4 [92.9, 97.9] \\
DBpedia 4 Nat. & Laya & 90.7 [86.7, 93.6] & 11.1 [7.9, 15.3] & 79.6 [74.6, 84.3] \\
DBpedia 4 Nat. & Jev & 96.1 [93.1, 97.8] & 2.1 [1.0, 4.6] & 93.9 [91.1, 96.4] \\
DBpedia 4 ID & Laya & 87.5 [83.1, 90.9] & 10.7 [7.6, 14.9] & 76.8 [71.4, 81.8] \\
DBpedia 4 ID & Jev & 97.9 [95.4, 99.0] & 2.5 [1.2, 5.1] & 95.4 [92.9, 97.9] \\
DBpedia 13 Nat. & Laya & 77.1 [71.9, 81.7] & 18.6 [14.5, 23.5] & 58.6 [52.5, 64.6] \\
DBpedia 13 Nat. & Jev & 80.7 [75.7, 84.9] & 1.4 [0.6, 3.6] & 79.3 [74.3, 84.3] \\
DBpedia 13 ID & Laya & 81.4 [76.5, 85.5] & 22.1 [17.7, 27.4] & 59.3 [53.6, 65.4] \\
DBpedia 13 ID & Jev & 87.9 [83.5, 91.2] & 0.7 [0.2, 2.6] & 87.1 [82.9, 90.7] \\
Emotion 2 Nat. & Laya & 76.7 [68.3, 83.3] & 45.8 [37.2, 54.7] & 30.8 [20.8, 42.5] \\
Emotion 2 Nat. & Jev & 85.8 [78.5, 91.0] & 55.8 [46.9, 64.4] & 30.0 [20.8, 40.0] \\
Emotion 2 ID & Laya & 83.3 [75.7, 88.9] & 45.8 [37.2, 54.7] & 37.5 [28.3, 46.7] \\
Emotion 2 ID & Jev & 79.2 [71.1, 85.5] & 44.2 [35.6, 53.1] & 35.0 [25.0, 45.8] \\
Emotion 4 Nat. & Laya & 69.2 [60.4, 76.7] & 46.7 [38.0, 55.6] & 22.5 [12.5, 32.5] \\
Emotion 4 Nat. & Jev & 68.3 [59.6, 76.0] & 35.8 [27.8, 44.7] & 32.5 [22.5, 43.3] \\
Emotion 4 ID & Laya & 65.8 [57.0, 73.7] & 45.8 [37.2, 54.7] & 20.0 [10.8, 29.2] \\
Emotion 4 ID & Jev & 58.3 [49.4, 66.8] & 28.3 [21.0, 37.0] & 30.0 [20.0, 40.8] \\
Emotion 5 Nat. & Laya & 68.3 [59.6, 76.0] & 45.8 [37.2, 54.7] & 22.5 [14.2, 31.7] \\
Emotion 5 Nat. & Jev & 58.3 [49.4, 66.8] & 32.5 [24.8, 41.3] & 25.8 [15.8, 35.8] \\
Emotion 5 ID & Laya & 59.2 [50.2, 67.5] & 43.3 [34.8, 52.3] & 15.8 [7.5, 24.2] \\
Emotion 5 ID & Jev & 50.0 [41.2, 58.8] & 20.8 [14.5, 28.9] & 29.2 [20.8, 38.3] \\
TREC 2 Nat. & Laya & 96.3 [90.9, 98.6] & 62.4 [53.0, 70.9] & 33.9 [24.8, 44.0] \\
TREC 2 Nat. & Jev & 71.6 [62.5, 79.2] & 1.8 [0.5, 6.4] & 69.7 [61.5, 78.0] \\
TREC 2 ID & Laya & 95.4 [89.7, 98.0] & 42.2 [33.4, 51.6] & 53.2 [43.1, 62.4] \\
TREC 2 ID & Jev & 71.6 [62.5, 79.2] & 2.8 [0.9, 7.8] & 68.8 [59.6, 77.1] \\
TREC 4 Nat. & Laya & 99.1 [95.0, 99.8] & 69.7 [60.5, 77.6] & 29.4 [20.2, 38.5] \\
TREC 4 Nat. & Jev & 43.1 [34.2, 52.5] & 0.9 [0.2, 5.0] & 42.2 [32.1, 51.4] \\
TREC 4 ID & Laya & 92.7 [86.2, 96.2] & 46.8 [37.7, 56.1] & 45.9 [35.8, 56.0] \\
TREC 4 ID & Jev & 43.1 [34.2, 52.5] & 1.8 [0.5, 6.4] & 41.3 [31.2, 51.4] \\
TREC 5 Nat. & Laya & 97.2 [92.2, 99.1] & 69.7 [60.5, 77.6] & 27.5 [18.3, 36.7] \\
TREC 5 Nat. & Jev & 24.8 [17.6, 33.6] & 0.0 [0.0, 3.4] & 24.8 [16.5, 33.0] \\
TREC 5 ID & Laya & 94.5 [88.5, 97.5] & 40.4 [31.6, 49.8] & 54.1 [45.0, 63.3] \\
TREC 5 ID & Jev & 26.6 [19.2, 35.6] & 0.0 [0.0, 3.4] & 26.6 [18.3, 34.9] \\
\bottomrule
\end{longtable}
\endgroup

\begingroup\footnotesize\setlength{\tabcolsep}{3pt}
\begin{longtable}{llrrr}
\caption{All score AUROCs with 2,000 paired-text bootstrap 95\% intervals. Max/entropy use plain requests; none uses requests with an added rejection option. Intervals are marginal.}\label{tab:all_auc}\\
\toprule
Group ($k$, names) & Model & $-\max p$ [95\%] & Entropy [95\%] & None [95\%]\\
\midrule\endfirsthead
\toprule
Group ($k$, names) & Model & $-\max p$ [95\%] & Entropy [95\%] & None [95\%]\\
\midrule\endhead
AG News 2 Nat. & Laya & 0.955 [0.911, 0.984] & 0.955 [0.911, 0.984] & 0.980 [0.958, 0.995] \\
AG News 2 Nat. & Jev & 0.804 [0.721, 0.873] & 0.804 [0.721, 0.873] & 0.936 [0.897, 0.971] \\
AG News 2 ID & Laya & 0.954 [0.912, 0.981] & 0.954 [0.912, 0.981] & 0.970 [0.941, 0.992] \\
AG News 2 ID & Jev & 0.864 [0.793, 0.922] & 0.864 [0.793, 0.922] & 0.948 [0.913, 0.978] \\
AG News 3 Nat. & Laya & 0.927 [0.872, 0.967] & 0.938 [0.891, 0.973] & 0.980 [0.961, 0.995] \\
AG News 3 Nat. & Jev & 0.777 [0.696, 0.848] & 0.781 [0.701, 0.849] & 0.931 [0.889, 0.965] \\
AG News 3 ID & Laya & 0.928 [0.877, 0.966] & 0.935 [0.886, 0.969] & 0.966 [0.937, 0.987] \\
AG News 3 ID & Jev & 0.820 [0.744, 0.881] & 0.823 [0.748, 0.884] & 0.949 [0.914, 0.977] \\
DBpedia 2 Nat. & Laya & 0.964 [0.947, 0.979] & 0.964 [0.947, 0.979] & 0.970 [0.957, 0.982] \\
DBpedia 2 Nat. & Jev & 0.974 [0.961, 0.987] & 0.974 [0.961, 0.987] & 0.999 [0.998, 1.000] \\
DBpedia 2 ID & Laya & 0.960 [0.942, 0.974] & 0.960 [0.942, 0.974] & 0.970 [0.956, 0.982] \\
DBpedia 2 ID & Jev & 0.988 [0.978, 0.996] & 0.988 [0.978, 0.996] & 0.999 [0.998, 1.000] \\
DBpedia 4 Nat. & Laya & 0.957 [0.941, 0.972] & 0.959 [0.944, 0.973] & 0.954 [0.938, 0.969] \\
DBpedia 4 Nat. & Jev & 0.928 [0.906, 0.947] & 0.928 [0.906, 0.948] & 0.996 [0.991, 0.999] \\
DBpedia 4 ID & Laya & 0.947 [0.929, 0.963] & 0.950 [0.932, 0.965] & 0.956 [0.941, 0.970] \\
DBpedia 4 ID & Jev & 0.958 [0.941, 0.975] & 0.958 [0.941, 0.975] & 0.996 [0.991, 0.999] \\
DBpedia 13 Nat. & Laya & 0.787 [0.747, 0.825] & 0.782 [0.744, 0.819] & 0.868 [0.841, 0.893] \\
DBpedia 13 Nat. & Jev & 0.976 [0.959, 0.989] & 0.976 [0.961, 0.989] & 0.992 [0.986, 0.998] \\
DBpedia 13 ID & Laya & 0.844 [0.810, 0.876] & 0.852 [0.819, 0.884] & 0.871 [0.846, 0.896] \\
DBpedia 13 ID & Jev & 0.975 [0.960, 0.987] & 0.975 [0.961, 0.987] & 0.995 [0.989, 0.999] \\
Emotion 2 Nat. & Laya & 0.621 [0.547, 0.698] & 0.621 [0.547, 0.698] & 0.747 [0.694, 0.804] \\
Emotion 2 Nat. & Jev & 0.620 [0.559, 0.683] & 0.620 [0.559, 0.683] & 0.740 [0.687, 0.795] \\
Emotion 2 ID & Laya & 0.656 [0.585, 0.727] & 0.656 [0.585, 0.727] & 0.767 [0.713, 0.822] \\
Emotion 2 ID & Jev & 0.614 [0.552, 0.679] & 0.614 [0.552, 0.679] & 0.736 [0.683, 0.790] \\
Emotion 4 Nat. & Laya & 0.610 [0.548, 0.676] & 0.620 [0.558, 0.685] & 0.655 [0.600, 0.716] \\
Emotion 4 Nat. & Jev & 0.533 [0.472, 0.594] & 0.538 [0.476, 0.598] & 0.705 [0.646, 0.763] \\
Emotion 4 ID & Laya & 0.604 [0.537, 0.672] & 0.616 [0.552, 0.682] & 0.646 [0.593, 0.700] \\
Emotion 4 ID & Jev & 0.543 [0.480, 0.607] & 0.551 [0.490, 0.613] & 0.696 [0.637, 0.755] \\
Emotion 5 Nat. & Laya & 0.635 [0.572, 0.696] & 0.644 [0.586, 0.702] & 0.676 [0.632, 0.723] \\
Emotion 5 Nat. & Jev & 0.517 [0.459, 0.572] & 0.517 [0.461, 0.570] & 0.698 [0.647, 0.750] \\
Emotion 5 ID & Laya & 0.615 [0.556, 0.675] & 0.632 [0.577, 0.688] & 0.656 [0.610, 0.704] \\
Emotion 5 ID & Jev & 0.520 [0.464, 0.576] & 0.525 [0.471, 0.578] & 0.698 [0.647, 0.750] \\
TREC 2 Nat. & Laya & 0.832 [0.771, 0.888] & 0.832 [0.771, 0.888] & 0.895 [0.854, 0.933] \\
TREC 2 Nat. & Jev & 0.813 [0.749, 0.872] & 0.813 [0.749, 0.872] & 0.945 [0.913, 0.972] \\
TREC 2 ID & Laya & 0.883 [0.832, 0.931] & 0.883 [0.832, 0.931] & 0.922 [0.885, 0.955] \\
TREC 2 ID & Jev & 0.829 [0.767, 0.888] & 0.829 [0.767, 0.888] & 0.941 [0.908, 0.971] \\
TREC 4 Nat. & Laya & 0.839 [0.782, 0.892] & 0.846 [0.791, 0.895] & 0.867 [0.823, 0.908] \\
TREC 4 Nat. & Jev & 0.686 [0.611, 0.761] & 0.695 [0.621, 0.769] & 0.903 [0.863, 0.939] \\
TREC 4 ID & Laya & 0.888 [0.843, 0.931] & 0.899 [0.859, 0.936] & 0.886 [0.846, 0.924] \\
TREC 4 ID & Jev & 0.707 [0.629, 0.779] & 0.717 [0.641, 0.787] & 0.888 [0.847, 0.927] \\
TREC 5 Nat. & Laya & 0.819 [0.758, 0.873] & 0.832 [0.776, 0.883] & 0.865 [0.820, 0.906] \\
TREC 5 Nat. & Jev & 0.630 [0.554, 0.704] & 0.634 [0.559, 0.708] & 0.846 [0.799, 0.892] \\
TREC 5 ID & Laya & 0.880 [0.833, 0.921] & 0.887 [0.843, 0.926] & 0.881 [0.838, 0.922] \\
TREC 5 ID & Jev & 0.652 [0.572, 0.726] & 0.658 [0.578, 0.731] & 0.843 [0.798, 0.886] \\
\bottomrule
\end{longtable}
\endgroup

\begingroup\footnotesize\setlength{\tabcolsep}{3pt}
\renewcommand{\arraystretch}{0.98}
\begin{longtable}{lllrr}
\caption{All calibration-only selected test operating points (percentages). Counts and Wilson intervals remain in frozen metrics; these are actual test rates, not a nominal 5\% guarantee.}\label{tab:all_calibration}\\
\toprule
Group & Model & Score & Test $D$ & Test $F$\\
\midrule\endfirsthead
\toprule
Group & Model & Score & Test $D$ & Test $F$\\
\midrule\endhead
AG News 2 Nat. & Laya & $-\max p$ & 86.3 & 8.8 \\
AG News 2 Nat. & Laya & Entropy & 86.3 & 8.8 \\
AG News 2 Nat. & Laya & None & 91.3 & 6.3 \\
AG News 2 Nat. & Jev & $-\max p$ & 41.3 & 5.0 \\
AG News 2 Nat. & Jev & Entropy & 41.3 & 5.0 \\
AG News 2 Nat. & Jev & None & 78.8 & 3.8 \\
AG News 2 ID & Laya & $-\max p$ & 83.8 & 8.8 \\
AG News 2 ID & Laya & Entropy & 83.8 & 8.8 \\
AG News 2 ID & Laya & None & 91.3 & 5.0 \\
AG News 2 ID & Jev & $-\max p$ & 65.0 & 8.8 \\
AG News 2 ID & Jev & Entropy & 65.0 & 8.8 \\
AG News 2 ID & Jev & None & 80.0 & 3.8 \\
AG News 3 Nat. & Laya & $-\max p$ & 75.0 & 11.3 \\
AG News 3 Nat. & Laya & Entropy & 71.3 & 6.3 \\
AG News 3 Nat. & Laya & None & 86.3 & 5.0 \\
AG News 3 Nat. & Jev & $-\max p$ & 40.0 & 7.5 \\
AG News 3 Nat. & Jev & Entropy & 43.8 & 7.5 \\
AG News 3 Nat. & Jev & None & 70.0 & 1.3 \\
AG News 3 ID & Laya & $-\max p$ & 73.8 & 8.8 \\
AG News 3 ID & Laya & Entropy & 73.8 & 3.8 \\
AG News 3 ID & Laya & None & 85.0 & 6.3 \\
AG News 3 ID & Jev & $-\max p$ & 43.8 & 6.3 \\
AG News 3 ID & Jev & Entropy & 46.3 & 5.0 \\
AG News 3 ID & Jev & None & 72.5 & 2.5 \\
DBpedia 2 Nat. & Laya & $-\max p$ & 81.4 & 2.5 \\
DBpedia 2 Nat. & Laya & Entropy & 81.4 & 2.5 \\
DBpedia 2 Nat. & Laya & None & 84.3 & 2.1 \\
DBpedia 2 Nat. & Jev & $-\max p$ & 95.0 & 1.1 \\
DBpedia 2 Nat. & Jev & Entropy & 95.0 & 1.1 \\
DBpedia 2 Nat. & Jev & None & 100.0 & 3.6 \\
DBpedia 2 ID & Laya & $-\max p$ & 72.1 & 2.5 \\
DBpedia 2 ID & Laya & Entropy & 72.1 & 2.5 \\
DBpedia 2 ID & Laya & None & 90.4 & 5.4 \\
DBpedia 2 ID & Jev & $-\max p$ & 97.9 & 2.1 \\
DBpedia 2 ID & Jev & Entropy & 97.9 & 2.1 \\
DBpedia 2 ID & Jev & None & 100.0 & 3.6 \\
DBpedia 4 Nat. & Laya & $-\max p$ & 75.7 & 3.6 \\
DBpedia 4 Nat. & Laya & Entropy & 82.5 & 6.4 \\
DBpedia 4 Nat. & Laya & None & 79.3 & 5.7 \\
DBpedia 4 Nat. & Jev & $-\max p$ & 85.7 & 0.7 \\
DBpedia 4 Nat. & Jev & Entropy & 85.7 & 0.7 \\
DBpedia 4 Nat. & Jev & None & 99.6 & 3.2 \\
DBpedia 4 ID & Laya & $-\max p$ & 70.0 & 3.6 \\
DBpedia 4 ID & Laya & Entropy & 74.3 & 5.0 \\
DBpedia 4 ID & Laya & None & 82.9 & 6.4 \\
DBpedia 4 ID & Jev & $-\max p$ & 87.9 & 0.4 \\
DBpedia 4 ID & Jev & Entropy & 87.9 & 0.4 \\
DBpedia 4 ID & Jev & None & 99.6 & 3.6 \\
DBpedia 13 Nat. & Laya & $-\max p$ & 27.9 & 3.2 \\
DBpedia 13 Nat. & Laya & Entropy & 32.9 & 3.6 \\
DBpedia 13 Nat. & Laya & None & 23.9 & 2.1 \\
DBpedia 13 Nat. & Jev & $-\max p$ & 85.7 & 1.4 \\
DBpedia 13 Nat. & Jev & Entropy & 87.5 & 1.8 \\
DBpedia 13 Nat. & Jev & None & 98.9 & 3.2 \\
DBpedia 13 ID & Laya & $-\max p$ & 27.5 & 3.2 \\
DBpedia 13 ID & Laya & Entropy & 37.1 & 3.2 \\
DBpedia 13 ID & Laya & None & 30.4 & 2.5 \\
DBpedia 13 ID & Jev & $-\max p$ & 78.2 & 1.4 \\
DBpedia 13 ID & Jev & Entropy & 81.4 & 1.4 \\
DBpedia 13 ID & Jev & None & 97.5 & 2.5 \\
Emotion 2 Nat. & Laya & $-\max p$ & 10.8 & 7.5 \\
Emotion 2 Nat. & Laya & Entropy & 10.8 & 7.5 \\
Emotion 2 Nat. & Laya & None & 29.2 & 9.2 \\
Emotion 2 Nat. & Jev & $-\max p$ & 0.8 & 0.8 \\
Emotion 2 Nat. & Jev & Entropy & 0.8 & 0.8 \\
Emotion 2 Nat. & Jev & None & 0.0 & 0.0 \\
Emotion 2 ID & Laya & $-\max p$ & 25.8 & 10.8 \\
Emotion 2 ID & Laya & Entropy & 25.8 & 10.8 \\
Emotion 2 ID & Laya & None & 35.8 & 10.0 \\
Emotion 2 ID & Jev & $-\max p$ & 6.7 & 2.5 \\
Emotion 2 ID & Jev & Entropy & 6.7 & 2.5 \\
Emotion 2 ID & Jev & None & 0.0 & 0.0 \\
Emotion 4 Nat. & Laya & $-\max p$ & 8.3 & 5.0 \\
Emotion 4 Nat. & Laya & Entropy & 5.8 & 4.2 \\
Emotion 4 Nat. & Laya & None & 21.7 & 9.2 \\
Emotion 4 Nat. & Jev & $-\max p$ & 5.0 & 4.2 \\
Emotion 4 Nat. & Jev & Entropy & 1.7 & 2.5 \\
Emotion 4 Nat. & Jev & None & 20.0 & 5.0 \\
Emotion 4 ID & Laya & $-\max p$ & 11.7 & 2.5 \\
Emotion 4 ID & Laya & Entropy & 7.5 & 3.3 \\
Emotion 4 ID & Laya & None & 20.0 & 7.5 \\
Emotion 4 ID & Jev & $-\max p$ & 7.5 & 5.8 \\
Emotion 4 ID & Jev & Entropy & 10.0 & 2.5 \\
Emotion 4 ID & Jev & None & 15.8 & 3.3 \\
Emotion 5 Nat. & Laya & $-\max p$ & 10.0 & 4.2 \\
Emotion 5 Nat. & Laya & Entropy & 13.3 & 6.7 \\
Emotion 5 Nat. & Laya & None & 18.3 & 5.8 \\
Emotion 5 Nat. & Jev & $-\max p$ & 4.2 & 2.5 \\
Emotion 5 Nat. & Jev & Entropy & 10.0 & 10.0 \\
Emotion 5 Nat. & Jev & None & 12.5 & 4.2 \\
Emotion 5 ID & Laya & $-\max p$ & 13.3 & 8.3 \\
Emotion 5 ID & Laya & Entropy & 16.7 & 8.3 \\
Emotion 5 ID & Laya & None & 15.8 & 5.8 \\
Emotion 5 ID & Jev & $-\max p$ & 5.8 & 3.3 \\
Emotion 5 ID & Jev & Entropy & 6.7 & 2.5 \\
Emotion 5 ID & Jev & None & 17.5 & 5.8 \\
TREC 2 Nat. & Laya & $-\max p$ & 52.3 & 10.1 \\
TREC 2 Nat. & Laya & Entropy & 52.3 & 10.1 \\
TREC 2 Nat. & Laya & None & 57.8 & 9.2 \\
TREC 2 Nat. & Jev & $-\max p$ & 27.5 & 3.7 \\
TREC 2 Nat. & Jev & Entropy & 27.5 & 3.7 \\
TREC 2 Nat. & Jev & None & 70.6 & 1.8 \\
TREC 2 ID & Laya & $-\max p$ & 55.0 & 5.5 \\
TREC 2 ID & Laya & Entropy & 55.0 & 5.5 \\
TREC 2 ID & Laya & None & 62.4 & 4.6 \\
TREC 2 ID & Jev & $-\max p$ & 29.4 & 4.6 \\
TREC 2 ID & Jev & Entropy & 29.4 & 4.6 \\
TREC 2 ID & Jev & None & 71.6 & 2.8 \\
TREC 4 Nat. & Laya & $-\max p$ & 39.4 & 5.5 \\
TREC 4 Nat. & Laya & Entropy & 45.0 & 7.3 \\
TREC 4 Nat. & Laya & None & 38.5 & 6.4 \\
TREC 4 Nat. & Jev & $-\max p$ & 10.1 & 0.9 \\
TREC 4 Nat. & Jev & Entropy & 15.6 & 0.9 \\
TREC 4 Nat. & Jev & None & 61.5 & 4.6 \\
TREC 4 ID & Laya & $-\max p$ & 67.9 & 12.8 \\
TREC 4 ID & Laya & Entropy & 79.8 & 14.7 \\
TREC 4 ID & Laya & None & 52.3 & 7.3 \\
TREC 4 ID & Jev & $-\max p$ & 13.8 & 3.7 \\
TREC 4 ID & Jev & Entropy & 22.0 & 2.8 \\
TREC 4 ID & Jev & None & 59.6 & 3.7 \\
TREC 5 Nat. & Laya & $-\max p$ & 53.2 & 6.4 \\
TREC 5 Nat. & Laya & Entropy & 58.7 & 6.4 \\
TREC 5 Nat. & Laya & None & 33.9 & 3.7 \\
TREC 5 Nat. & Jev & $-\max p$ & 13.8 & 8.3 \\
TREC 5 Nat. & Jev & Entropy & 8.3 & 0.9 \\
TREC 5 Nat. & Jev & None & 45.0 & 1.8 \\
TREC 5 ID & Laya & $-\max p$ & 51.4 & 7.3 \\
TREC 5 ID & Laya & Entropy & 51.4 & 6.4 \\
TREC 5 ID & Laya & None & 46.8 & 7.3 \\
TREC 5 ID & Jev & $-\max p$ & 10.1 & 1.8 \\
TREC 5 ID & Jev & Entropy & 13.8 & 2.8 \\
TREC 5 ID & Jev & None & 43.1 & 0.9 \\
\bottomrule
\end{longtable}
\endgroup

\section{Related evaluation settings}
\paragraph{Direct Jev rejection study.} Zhong et al. evaluate paired present/absent numerical menus, compare decision interfaces, include a generative control and calibrate rejection on separate development problems \cite{zhongWhenRightAnswer2026}. Pairing, local calibration and an LLM comparison therefore already have direct precedents. Our benchmark adds shared natural-language task measurements, fixed-budget source-policy transfer and distinct artificial/retrieved/OOS populations. It does not claim priority for paired omission or threshold calibration.
\paragraph{Missing answers and rejection.} Tam et al. study none-of-the-above options in LLM multiple-choice evaluation \cite{tamNoneLessRight2025}; Wang et al. diagnose absent answers in video-language tasks \cite{wangWhenNoAnswer2026}; Feng et al. consider none-of-the-above uncertainty in dialog retrieval \cite{fengNoneMeasureUncertainty2020}. These establish answer absence as a relevant evaluation target. Our protocol pairs reference-label absence with present controls at the same ordinary candidate count and measures false rejection, classification competence, and calibration-selected operating points together.

Open-set recognition highlights the relationship between rejection and closed-set classification \cite{vazeOpenSetRecognitionGood2022}. Intent work evaluates unsupported or out-of-distribution queries \cite{zhanOutofDistributionIntentDetection2021,larsonEvaluationDatasetIntent2019}. Our omitted-label conditions are not naturally novel or OOS examples; their reference labels are known but removed from the request. Selective prediction instead concerns the error risk of accepted outputs \cite{wangLECLinearExpectation2026}. We distinguish candidate coverage from the acceptance fraction and make no risk-control guarantee from empirical calibration constraints.

\paragraph{Typed decision interfaces.} Sun et al. change option-name/definition associations \cite{sunTypeSafeNotErrorFree2026}; Joy studies coherence across hierarchical formulations in Jev and Laya \cite{joySystemOneDecisions2026}; JevOut examines decision flips induced by surrounding context \cite{xuJevOutNaturalContext2026}. We retain texts and retained rubrics, vary membership at matched size, and compare natural/neutral keys. Naming sensitivity is a supporting diagnostic, not a new phenomenon claimed here. The present evaluation complements those studies with operational missing-answer and false-rejection measurements, keeping scoring and native choices separate.

\clearpage
\bibliographystyle{unsrtnat}
\bibliography{references}
\end{document}